\documentclass[11pt]{article}
\PassOptionsToPackage{hyphens}{url}

\usepackage[preprint]{acl}

\usepackage{times}
\usepackage{latexsym}
\usepackage[T1]{fontenc}
\usepackage[utf8]{inputenc}
\usepackage{microtype}
\IfFileExists{inconsolata.sty}{\usepackage{inconsolata}}{}

\usepackage{amsmath, amssymb}
\usepackage{algorithm}
\usepackage{algpseudocode}
\usepackage{booktabs}
\usepackage{array}
\usepackage{graphicx}
\usepackage{xcolor}
\usepackage{pgfplots}
\pgfplotsset{compat=1.17}
\usepgfplotslibrary{groupplots}

\newcommand{\Ndraft}{N_{\mathrm{draft}}}
\newcommand{\laccept}{\bar{\ell}}          % mean accepted draft tokens per round
\newcommand{\Lbar}{\bar{L}}                % mean completion length
\newcommand{\crel}{c_{\mathrm{rel}}}       % drafter step cost relative to one target step
\newcommand{\TV}{\mathrm{TV}}
\newcommand{\KL}{D_{\mathrm{KL}}}
\newcommand{\pirej}{\pi^{\mathrm{rej}}}
\newcommand{\pires}{\pi^{\mathrm{res}}}
\newcommand{\pibonus}{\pi^{\mathrm{bonus}}}
\DeclareMathOperator{\Ber}{Ber}
\DeclareMathOperator{\normop}{norm}
\DeclareMathOperator{\Div}{Div}
\DeclareMathOperator{\softmax}{softmax}

\algnewcommand{\LineComment}[1]{\State \(\triangleright\) \textit{#1}}

\newcommand{\finding}[1]{\textbf{#1}\enspace}

\definecolor{Mmentoreddec}{HTML}{0072B2}
\definecolor{Mcactus}{HTML}{D55E00}
\definecolor{Mspeccascopt}{HTML}{009E73}
\definecolor{Mrfuzzy}{HTML}{CC79A7}
\definecolor{Mspeccasctok}{HTML}{E69F00}
\definecolor{Deagle}{HTML}{D55E00}
\definecolor{Dmid}{HTML}{009E73}
\definecolor{Dlm}{HTML}{0072B2}

\title{Acceptance Is Not Speed: Length Inflation in Training-Free Lossy Speculative Decoding}

\author{
  Chiatzen Wang \quad Bill Xu \quad Haochen Gu \quad Hongyang Zhang \\
  University of Waterloo \\
  \texttt{chiatzenw@outlook.com}, \texttt{bill.xu1@uwaterloo.ca} \\
  \texttt{haochen.gu@uwaterloo.ca}, \texttt{hongyang.zhang@uwaterloo.ca}
}

\begin{document}
\maketitle

\begin{abstract}
Speculative decoding typically characterizes speedup by draft tokens accepted per verification round. Lossless verification preserves the target model's output distribution and thus expected completion length, making this metric a meaningful efficiency proxy. Relaxed (lossy) verification accepts tokens that lossless verification would reject, potentially altering that distribution. We identify a length inflation phenomenon: relaxed verification lengthens completions, primarily by extending reasoning traces, largely offsetting apparent gains in accepted tokens per round. We evaluate existing training-free relaxed verification rules on twelve target-drafter pairs against lossless verification on identical prompts, measuring completion length, verifier rounds, wall-clock time, and task accuracy. At each rule's most relaxed evaluated setting, relaxed verification on average accepts $1.34\times$ as many tokens per round but produces $1.54\times$ longer completions. For rules with higher acceptance lengths, tokens accepted only under relaxed verification tend to have low target-model probability, contributing to length inflation. For rules that reduce verifier rounds, accuracy typically drops substantially. Further analysis traces some failures to continued reasoning that exhausts the budget needed to deliver a final answer. Overall, accepted tokens per round alone is an unreliable measure of decoding speed under relaxed verification. Completion length and verifier rounds must also be reported to assess actual efficiency.
\end{abstract}

\section{Introduction}
\label{sec:intro}

Large language models generate text autoregressively: every new token requires a full forward pass over the prefix. Speculative decoding accelerates this loop by separating proposal from verification. A cheap draft model proposes several tokens ahead, and the target model scores all of them in one forward pass, committing the longest accepted prefix \citep{stern2018blockwise,leviathan2023fast,chen2023accelerating}. The technique is now standard in production inference stacks, with learned drafters such as the EAGLE family \citep{li2024eagle,li2024eagle2,li2025eagle3} shipped in vLLM \citep{kwon2023vllm}, SGLang \citep{zheng2024sglang}, and TensorRT-LLM \citep{nvidia2023tensorrtllm}. We call such drafters \emph{dedicated}: a dedicated drafter computes its proposals from the target's own hidden states, so it is trained for one target and cannot run without it.\footnote{Formally, a dedicated drafter samples from $q(\cdot \mid x_{<t}, h_{<t})$, where $h_{<t}$ are the target's hidden states on the prefix $x_{<t}$, and a standalone drafter from $q(\cdot \mid x_{<t})$.} For example, the EAGLE-3 head of Llama-3.1-8B-Instruct is a single transformer layer that drafts from features of a low, a middle, and a high layer of the target. A \emph{standalone} drafter is instead a complete small language model that drafts from the tokens alone and shares the target's tokenizer, such as Llama-3.2-1B-Instruct for the same target.

In lossless speculative decoding the acceptance rule preserves the target distribution exactly, so the speedup is capped by how closely the drafter tracks the target \citep{leviathan2023fast}. Because the output distribution is unchanged, completion length is unchanged too, and a gain in accepted tokens per round is a gain in speed.

Lossy speculative decoding relaxes the acceptance rule to admit more draft tokens, trading exactness of the output distribution for speed. It also gives up the guarantee on length: a relaxed rule decodes a different distribution, and its completions can be longer or shorter than the target's. A growing family of training-free methods does this by replacing the target distribution inside verification with a relaxed distribution $\pi$ controlled by a relaxation parameter $\alpha$ \citep{xia2026practical,li2026arcdecode}. These methods report higher acceptance and throughput at what their evaluations describe as negligible quality cost.

Those gains, however, are measured almost entirely in accepted tokens per round and tokens per second \citep{wang2026revisiting}. A relaxed rule changes which tokens are committed, and therefore how long its completions run and whether they terminate cleanly. If a method accepts more tokens per round but also inflates completion length, the end-to-end gain shrinks and can invert. We call this \emph{length inflation} under relaxed verification. The effect is invisible to both metrics. This paper measures it directly (Sections~\ref{sec:length} and~\ref{sec:rounds}), characterizes it across rules, tasks, targets, drafters, and settings (Sections~\ref{sec:accuracy} and~\ref{sec:analysis}), and traces it to the low-probability tokens that relaxation commits (Section~\ref{sec:why}). The effect matters for reasoning models, whose thinking traces run to thousands of tokens. The added length buys no accuracy: forcing the reasoning trace to end near the generation budget recovers far more correct answers than it loses (Section~\ref{sec:buys}).

\finding{Differences with prior work.}
Prior work has not sufficiently characterized the extent and consequences of length inflation under relaxed verification. \citet{xia2026practical} discuss longer, rambling responses only in weak-drafter settings. \citet{wang2026revisiting} study quality degradation but do not measure completion length. By systematically treating length inflation as a central measurement of relaxed verification, we identify it as a consistent phenomenon across lossy speculative decoding methods, whose final outputs are longer than their lossless counterparts. We further investigate what the additional generation accomplishes and whether its consequences are recoverable. These findings support a concrete explanation for some unsuccessful responses: continued reasoning exhausts the budget needed to deliver a final answer. In summary, we uncover a new prevalent phenomenon in lossy speculative decoding and identify its cause.

We evaluate five training-free relaxed verification rules, namely CACTUS~\citep{hao2026cactus}, mentored decoding~\citep{tranthien2023mentored}, reducible fuzzy speculative decoding~\citep{holsman2025fuzzy}, and the two speculative cascade rules~\citep{narasimhan2025cascades}, implemented as published, on six benchmarks covering math, code, chat, and long context, with twelve target-drafter pairs that combine four targets (GPT-OSS-20B, Qwen3-8B, Llama-3.1-8B-Instruct, and DeepSeek-R1-Distill-Llama-8B) with eight dedicated drafters and three standalone language models (360 cells in all). Our contributions are:
\begin{enumerate}
    \item a token-efficiency accounting for relaxed speculative decoding: completion length, verifier rounds, the condition under which a relaxed rule saves decoding time, and the largest length inflation any relaxation of a given drafter can absorb (Section~\ref{sec:framework}), with a protocol that compares methods at matched accepted length rather than matched $\alpha$;
    \item a cross-method, benchmark, target, and drafter measurement over 360 cells (Section~\ref{sec:experiments}), with paired bootstrap intervals for every cell, showing that relaxation inflates completion length, that the inflation cancels most of the promised reduction in verifier rounds and in measured time, that this holds at gentle settings as well as loose ones, and that the accuracy cost is large at loose settings and task dependent;
    \item the identification of where the extra length comes from: relaxation admits draft tokens the target rates as unlikely, and the extra text goes into the reasoning trace where the model has one (Section~\ref{sec:why});
    \item evidence that the added length is waste: on grade-school math it follows a correct answer the model has already written, on competition math it delays the answer, and forcing the reasoning trace to end near the generation budget recovers far more correct answers (Section~\ref{sec:buys}).
\end{enumerate}

\section{Relaxed Speculative Decoding}
\label{sec:methods}

This section presents lossless and lossy speculative decoding in one framework and states every training-free method as a choice of relaxed target distributions in it; following \citet{xia2026practical} we also call the two cases strict and relaxed. Appendix~\ref{app:methods} gives each method's rule and our implementation.

\subsection{Strict speculative decoding}
\label{sec:strict}

Let $p$ be the target model to accelerate and $q$ a cheaper drafter, with next-token distributions $p_t(\cdot) = p(\cdot \mid x_{<t})$ and $q_t(\cdot) = q(\cdot \mid x_{<t})$ over a vocabulary of size $V$, and $x$ a drafted token. In one round of speculative decoding (Algorithm~\ref{alg:specdec} in Appendix~\ref{app:methods}, with $\pi = p$), the drafter samples a block of $\Ndraft$ tokens and the target scores all of them, plus the position after the block, in a single forward pass. Each draft token is then accepted with probability $\min\{1, p(x)/q(x)\}$; at the first rejection the token is replaced by a sample from the residual distribution $\normop(\max\{0, p - q\})$ and the round ends. If every draft token is accepted, a bonus token is sampled from $p$ at the position after the block. One target forward pass thus commits between one and $\Ndraft + 1$ tokens, and the output is distributed exactly as $p$ \citep{leviathan2023fast,chen2023accelerating}.

Let $C_{\mathrm{AR}}$, $C_{\mathrm{drafter}}$, and $C_{\mathrm{verify}}$ be the wall-clock times of one decoding step of the target $p$, one decoding step of the drafter $q$, and one verification pass over the $\Ndraft + 1$ positions, and let $\laccept$ denote the mean number of draft tokens accepted per round. A round costs $\Ndraft\, C_{\mathrm{drafter}} + C_{\mathrm{verify}}$ and commits $\laccept + 1$ tokens on average. Under memory-bound conditions $C_{\mathrm{verify}} \approx C_{\mathrm{AR}}$, so with $\crel = C_{\mathrm{drafter}} / C_{\mathrm{AR}}$ the speedup is
\begin{equation}
    S \approx \frac{(\laccept + 1)\, C_{\mathrm{AR}}}{\Ndraft\, C_{\mathrm{drafter}} + C_{\mathrm{verify}}}
      \approx \frac{\laccept + 1}{1 + \Ndraft\, \crel}.
    \label{eq:speedup}
\end{equation}
The accepted length is bounded by drafter--target agreement: a draft token drawn from $q$ is accepted with probability
\begin{equation}
    \textstyle\sum_{v} \min\{p(v), q(v)\} = 1 - \TV(p, q),
    \label{eq:tv}
\end{equation}
so strict verification cannot accept more than the two distributions share \citep{leviathan2023fast}. Every relaxed method below is an attempt to lift this cap.

\subsection{Relaxed speculative decoding}
\label{sec:framework}

A relaxed method keeps the round of Section~\ref{sec:strict} intact and replaces the target distribution $p$ at up to three points with relaxed target distributions $\pi$: $\pirej$ in the acceptance distribution, so that a draft token is accepted with probability $\min\{1, \pirej(x)/q(x)\}$; $\pires$ in the residual sampled after a rejection; and $\pibonus$ at the bonus position. Each is a function $\pi[q, p, x, \alpha](\cdot)$ of the two distributions, the drafted token, and a method-specific relaxation parameter $\alpha$. Strict speculative decoding is the special case $\pi = p$ at all three points. The parameter $\alpha$ means something different in every method (a divergence budget, a divergence threshold, a deferral cost, a mixture weight), so values of $\alpha$ are never comparable across methods; Section~\ref{sec:experiments} therefore compares methods at matched $\laccept$ rather than matched $\alpha$.

\paragraph{Scope.}
We restrict attention to training-free rules that relax verification per token, i.e.\ methods fully specified by a choice of $(\pirej, \pires, \pibonus)$; rules that train a component, such as a judge head \citep{bachmann2025judge,garipov2025autojudge} or a learned ensemble weight \citep{wang2026diversed}, expose no $\alpha$ to sweep.

\paragraph{Length-corrected speedup.}
Because a relaxed rule samples from $\pi$ rather than $p$, it can change the length of the generated response, and Equation~\eqref{eq:speedup} no longer measures end-to-end speed. We can correct for this with
\begin{equation}
    S_{\mathrm{relax}} \approx \frac{\Lbar_{\mathrm{AR}}}{\Lbar_{\mathrm{relax}}} \cdot S,
    \label{eq:speedup-relax}
\end{equation}
where $\Lbar$ denotes mean completion length and $S$ is the acceptance-based speedup of Equation~\eqref{eq:speedup}. This study measures that factor directly: a method wins end to end only if its acceptance gain exceeds its length inflation.

\paragraph{Verifier rounds.}
The same trade-off can be measured without a hardware model. A completion of $L$ tokens generated in $R$ verifier rounds (target forward passes) commits $\laccept + 1$ tokens per round on average, so
\begin{equation}
    R \approx \frac{L}{\laccept + 1}.
    \label{eq:rounds}
\end{equation}
Every round costs $\Ndraft$ drafter steps and one verification pass for every rule in the sweep, so at fixed $\Ndraft$ and drafter the decoding times of two rules are in the ratio of their round counts. Writing $\lambda = \Lbar_{\mathrm{relax}} / \Lbar_{\mathrm{strict}}$ for the length inflation relative to strict verification,
\begin{equation}
    \frac{R_{\mathrm{relax}}}{R_{\mathrm{strict}}} \approx \lambda \cdot \frac{\laccept_{\mathrm{strict}} + 1}{\laccept_{\mathrm{relax}} + 1},
    \label{eq:netwin}
\end{equation}
and a relaxed rule reduces decoding time only if $(\laccept_{\mathrm{relax}} + 1) / (\laccept_{\mathrm{strict}} + 1) > \lambda$. The gain on the left is bounded by the draft length: a round accepts at most $\Ndraft$ draft tokens, so no relaxation of a given drafter can beat $\lambda = (\Ndraft + 1)/(\laccept_{\mathrm{strict}} + 1)$. Section~\ref{sec:experiments} reports $\Lbar$, $R$, measured time, and task accuracy directly, at matched $\laccept$.

\subsection{Training-free methods}
\label{sec:trainingfree}

Table~\ref{tab:taxonomy} states every method as a choice of $(\pirej, \pires, \pibonus)$. The five swept rules fall into two groups: divergence-budgeted rules that boost the drafted token while bounding the distance from $p$ (CACTUS, mentored decoding), and switch rules that trust the drafter outright when a per-position test passes (fuzzy speculative decoding, speculative cascades).

The cascade token rule (\texttt{spec\_casc\_tok}) is not strict verification even at $\alpha = 0$: draft tokens inside a trusted top set of $p$ are accepted outright, and tokens outside it are judged against a scaled-down copy of $p$, so it is the only rule in the sweep that can reject what lossless verification would accept.

\section{Experimental Evaluation}
\label{sec:experiments}

We measure what relaxed verification does to the generated text: five training-free rules, \texttt{mentored\_dec}, \texttt{cactus}, \texttt{spec\_casc\_opt}, \texttt{r\_fuzzy}, and \texttt{spec\_casc\_tok}, run on six datasets with twelve target-drafter pairs, each compared with lossless (strict) speculative decoding on the same pair. One (pair, dataset, rule) combination is a cell, and there are 360. For every cell we report completion length, verifier rounds (target forward passes), measured time, and task accuracy.

\subsection{Setup}
\label{sec:setup}

\paragraph{Targets and drafters.}
We use four targets in twelve target-drafter pairs (Table~\ref{tab:pairs}): GPT-OSS-20B~\citep{openai2025gptoss} with two EAGLE-3 heads~\citep{li2025eagle3}; Qwen3-8B~\citep{yang2025qwen3} with an EAGLE-3 head, P-EAGLE~\citep{hui2026peagle}, DSpark~\citep{cheng2026dspark}, and the standalone Qwen3-0.6B and Qwen3-1.7B; Llama-3.1-8B-Instruct~\citep{grattafiori2024llama3} with EAGLE-3, EAGLE-1~\citep{li2024eagle}, and the standalone Llama-3.2-1B-Instruct; and DeepSeek-R1-Distill-Llama-8B~\citep{deepseek2025r1} with EAGLE-3 and Llama-3.2-1B-Instruct (checkpoints in Appendix~\ref{app:pairs}). The EAGLE-1, EAGLE-3, P-EAGLE, and DSpark heads are the eight dedicated drafters; Qwen3-0.6B, Qwen3-1.7B, and Llama-3.2-1B-Instruct are the three standalone ones. Two pairs, GPT-OSS-20B with the nebius EAGLE-3 head and Qwen3-8B with the RedHatAI one,\footnote{\texttt{nebius/EAGLE3-gpt-oss-20b} and \texttt{RedHatAI/Qwen3-8B-speculator.eagle3} on the Hugging Face Hub.} are measured in more depth (more seeds, timing on one server, run-log analyses, and the controls of Appendices~\ref{app:results}, \ref{app:seeds}, and~\ref{app:speedbench}); we call them the primary pairs. Every pair runs on vLLM 0.26.0 on a single NVIDIA H100 with $\Ndraft = 6$ (within $10\%$ of the fastest lossless draft length; Appendix~\ref{app:seeds}), sampled drafts, temperature $1.0$, top-$p$ $1.0$, and seed $0$; the bundled sampling defaults of Qwen3-8B and the Llama models are overridden (Appendices~\ref{app:seeds} and~\ref{app:pairs}).

\paragraph{Implementation.}
vLLM exposes no relaxed acceptance rule, so each method is a patch to its rejection sampler that implements the formulas of Table~\ref{tab:taxonomy}, applied one at a time. Each patch ships with tests that check the formula against closed-form values.\footnote{The EAGLE-3 head of Qwen3-8B, the EAGLE heads of the two Llama-based targets, and DSpark run through a second sampler path in vLLM, on which \texttt{cactus} and \texttt{spec\_casc\_tok} modify the acceptance test only and the residual distribution is the unmodified target's; the other pairs apply the full rules.} Code, patches, prompts, and per-run outputs are released.\footnote{Anonymized repository: \url{https://anonymous.4open.science/r/lossy-token-eff}.}

\paragraph{Datasets.}
Six datasets cover four regimes: mathematics (GSM8K~\citep{cobbe2021gsm8k}, AIME24~\citep{maa2024aime}), code generation (HumanEval~\citep{chen2021humaneval}, LiveCodeBench~\citep{jain2025livecodebench}), open-ended chat (MT-Bench~\citep{zheng2023mtbench}), and long-context multiple-choice question answering (LongBench-v2~\citep{bai2025longbench2}). GSM8K, AIME24, and LongBench-v2 are graded by exact match on the final answer; HumanEval and LiveCodeBench by executing the generated code against the problem's tests. A run that never produces an answer is scored wrong. MT-Bench is scored by an LLM judge (Table~\ref{tab:judge}). Case counts, generation budgets, and lossless references are listed in Tables~\ref{tab:strict} and~\ref{tab:strict-all}. A run that ends at its budget is counted as cut off.

\paragraph{Metrics.}
We report five quantities: $\laccept$, the mean accepted length (draft tokens accepted per round); $\Lbar$, the mean completion length in tokens; $R$, the mean number of verifier rounds; the mean per-request wall-clock time, measured at the client at batch size 1 and excluding server start; and accuracy, the pass rate over the dataset's cases (for MT-Bench, the mean judge score). Each case is run once per arm. Every ratio to lossless is computed on the same cases for both arms, and its uncertainty is a 95\% paired bootstrap interval over cases (10\,000 resamples). A cell on a graded dataset (all but MT-Bench) loses accuracy when it falls two or more points below its lossless reference.

\subsection{Matching on mean accepted length}
\label{sec:protocol}

Because $\alpha$ is not comparable across methods (Section~\ref{sec:framework}), we compare methods at matched $\laccept$. A short pilot runs each method's four $\alpha$ values (Table~\ref{tab:taxonomy}) on three cases per dataset; we then fix three target values of $\laccept$ and run the full dataset at the $\alpha$ closest to each. The five methods never share a reachable $\laccept$, so the match is approximate. The four pairs with a standalone drafter and the P-EAGLE pair run each rule at its loosest $\alpha$ only. The tables report each method at its most relaxed setting; Sections~\ref{sec:length} and~\ref{sec:analysis} check that the conclusions do not depend on this choice.

\subsection{Completion length}
\label{sec:length}

Relaxation lengthens completions in 248 of the 360 cells (Tables~\ref{tab:length} and~\ref{tab:pairs}): $\lambda$ has median $1.08$ and mean $1.54$, with the 95\% interval above $1$ in 172 cells and below $1$ in 67. Its median varies more with the rule ($0.97$ for \texttt{spec\_casc\_tok} to $1.39$ for \texttt{cactus}) and the drafter ($0.94$ for standalone drafters to $1.23$ for EAGLE heads) than with the task ($1.04$ on GSM8K to $1.14$ on HumanEval), and cut-off runs make it a lower bound (Section~\ref{sec:why}).

The inflation persists at each rule's gentlest setting (on the seven pairs run with the matched protocol, median $\lambda$ $1.07$ against $1.22$ at the loosest, and more rounds than lossless in 97 cells against 94; Table~\ref{tab:gentle}) and across seeds (Appendix~\ref{app:seeds}). At Qwen3-8B's recommended sampler (temperature $0.6$), \texttt{mentored\_dec}, \texttt{cactus}, and \texttt{spec\_casc\_tok} stop inflating on GSM8K and LiveCodeBench, but \texttt{spec\_casc\_opt} still inflates by $1.5\times$ (Appendix~\ref{app:seeds}).

\begin{table*}[t]
\centering\scriptsize
\renewcommand{\arraystretch}{0.9}
\setlength{\tabcolsep}{2.6pt}
\begin{tabular}{@{}l ccccc ccccc@{}}
\toprule
Dataset & mentored & cactus & casc-opt & r-fuzzy & casc-tok & mentored & cactus & casc-opt & r-fuzzy & casc-tok \\
\midrule
 & \multicolumn{5}{c}{GPT-OSS-20B + EAGLE-3 (nebius)} & \multicolumn{5}{c}{Qwen3-8B + EAGLE-3 (RedHatAI)} \\
\cmidrule(lr){2-6}\cmidrule(lr){7-11}
GSM8K & 1.16 (.93) & 1.61 (.91) & 1.46 (.94) & 1.70 (.87) & 0.96$^{\circ}$ (.97) & 1.03$^{\circ}$ (.77) & 1.04$^{\circ}$ (.77) & 1.33 (.49) & 0.95$^{\circ}$ (.55) & 0.99$^{\circ}$ (.79) \\
AIME24 & 1.69 (.63) & 1.78 (.60) & 2.48 (.37) & 2.13 (.50) & 1.48 (.70) & 1.01$^{\circ}$ (.73) & 1.26 (.23) & 1.42 (.30) & 1.29 (.40) & 0.98$^{\circ}$ (.70) \\
HumanEval & 1.30 (.96) & 1.79 (.87) & 1.73 (.82) & 1.64 (.63) & 1.05$^{\circ}$ (.94) & 1.08 (.85) & 1.14 (.75) & 1.72 (.49) & 1.65 (.16) & 1.08 (.85) \\
LiveCodeBench & 1.25 (.84) & 1.54 (.61) & 1.64 (.49) & 1.65 (.34) & 1.16 (.88) & 1.05 (.63) & 1.15 (.44) & 1.27 (.33) & 1.07$^{\circ}$ (.02) & 1.00$^{\circ}$ (.71) \\
MT-Bench & 1.10$^{\circ}$ (6.41) & 1.18 (4.51) & 1.09$^{\circ}$ (5.74) & 1.20 (4.58) & 0.96$^{\circ}$ (7.49) & 1.04$^{\circ}$ (6.40) & 1.16 (2.99) & 1.22 (3.95) & 1.12 (3.05) & 1.04$^{\circ}$ (7.21) \\
LongBench-v2 & 1.59 (.51) & 1.77 (.36) & 1.79 (.47) & 1.69 (.40) & 1.13$^{\circ}$ (.53) & 1.06$^{\circ}$ (.51) & 2.10 (.29) & 1.16 (.37) & 1.08$^{\circ}$ (.44) & 1.00$^{\circ}$ (.49) \\
\bottomrule
\end{tabular}
\caption{Length inflation on the primary pairs (Section~\ref{sec:setup}) at each method's most relaxed tested setting: completion length relative to lossless ($\lambda$; $>1$ is inflation), with the accuracy at that setting in parentheses (MT-Bench: judge score out of 10). $^{\circ}$: the 95\% interval for $\lambda$ includes $1$ (Table~\ref{tab:cells}). All twelve pairs: Table~\ref{tab:length-all}.}
\label{tab:length}
\end{table*}

\subsection{Verifier rounds and time}
\label{sec:rounds}

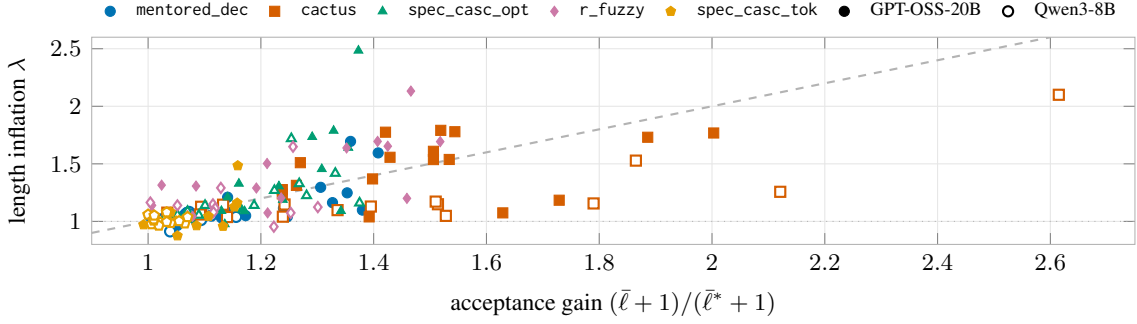
\begin{figure*}[t]
\centering
\nolinenumbers
\begin{tikzpicture}
\begin{axis}[width=0.96\textwidth, height=0.27\textwidth, tick label style={font=\footnotesize}, label style={font=\footnotesize}, grid=major, grid style={gray!20}, axis line style={gray!60}, xlabel={acceptance gain $(\laccept+1)/(\laccept^{*}+1)$}, ylabel={length inflation $\lambda$}, xmin=0.9, xmax=2.75, ymin=0.8, ymax=2.6, legend style={at={(0.5,1.03)}, anchor=south, font=\scriptsize, draw=none, fill=none, column sep=6pt}, legend columns=7, legend cell align=left]
\addplot[gray!60, dashed, thick, domain=0.9:2.6, forget plot] {x};
\addplot[gray!60, dotted, forget plot] coordinates {(0.9,1) (2.75,1)};
\addplot[only marks, mark=*, mark size=1.9pt, mark options={fill=Mmentoreddec, draw=Mmentoreddec}] coordinates {(1.086,0.996) (1.173,1.050) (1.327,1.163) (1.068,1.076) (1.141,1.210) (1.359,1.695) (1.052,0.965) (1.130,1.031) (1.306,1.296) (1.070,1.080) (1.141,1.082) (1.353,1.248) (1.050,0.999) (1.247,1.041) (1.380,1.098) (1.065,1.062) (1.240,1.114) (1.408,1.595)};
\addlegendentry{\texttt{mentored\_dec}}
\addplot[forget plot, only marks, mark=*, mark size=1.9pt, mark options={fill=white, draw=Mmentoreddec, thick}] coordinates {(1.017,0.991) (1.093,1.026) (1.039,0.913) (1.094,1.012) (1.004,1.046) (1.073,1.082) (1.019,1.018) (1.112,1.050) (1.025,1.017) (1.156,1.039) (1.012,0.988) (1.074,1.062)};
\addplot[only marks, mark=square*, mark size=1.9pt, mark options={fill=Mcactus, draw=Mcactus}] coordinates {(1.238,1.276) (1.429,1.557) (1.506,1.609) (1.270,1.510) (1.421,1.775) (1.544,1.779) (1.238,1.239) (1.519,1.791) (1.263,1.313) (1.398,1.369) (1.534,1.538) (1.392,1.039) (1.629,1.074) (1.729,1.184) (1.506,1.537) (1.886,1.731) (2.003,1.768)};
\addlegendentry{\texttt{cactus}}
\addplot[forget plot, only marks, mark=square*, mark size=1.9pt, mark options={fill=white, draw=Mcactus, thick}] coordinates {(1.043,1.016) (1.140,1.038) (1.242,1.147) (1.528,1.048) (2.121,1.257) (1.034,1.076) (1.094,1.127) (1.133,1.145) (1.092,1.060) (1.336,1.098) (1.514,1.149) (1.239,1.041) (1.395,1.129) (1.790,1.156) (1.510,1.172) (1.865,1.528) (2.615,2.100)};
\addplot[only marks, mark=triangle*, mark size=1.9pt, mark options={fill=Mspeccascopt, draw=Mspeccascopt}] coordinates {(1.239,1.188) (1.171,1.089) (1.308,1.455) (1.161,1.328) (1.373,2.484) (1.130,1.091) (1.291,1.734) (1.166,1.103) (1.355,1.640) (1.136,0.976) (1.342,1.091) (1.232,1.305) (1.141,1.209) (1.329,1.788)};
\addlegendentry{\texttt{spec\_casc\_opt}}
\addplot[forget plot, only marks, mark=triangle*, mark size=1.9pt, mark options={fill=white, draw=Mspeccascopt, thick}] coordinates {(1.090,1.054) (1.043,1.003) (1.268,1.328) (1.064,1.027) (1.332,1.418) (1.101,1.139) (1.031,1.043) (1.254,1.719) (1.052,1.055) (1.224,1.269) (1.188,1.139) (1.074,1.019) (1.281,1.224) (1.070,1.087) (1.375,1.160)};
\addplot[only marks, mark=diamond*, mark size=1.9pt, mark options={fill=Mrfuzzy, draw=Mrfuzzy}] coordinates {(1.111,1.054) (1.236,1.204) (1.407,1.695) (1.024,1.315) (1.085,1.307) (1.466,2.132) (1.016,1.022) (1.070,1.050) (1.352,1.638) (1.016,1.010) (1.192,1.290) (1.425,1.653) (1.079,1.044) (1.212,1.074) (1.459,1.200) (1.007,1.139) (1.211,1.503) (1.518,1.693)};
\addlegendentry{\texttt{r\_fuzzy}}
\addplot[forget plot, only marks, mark=diamond*, mark size=1.9pt, mark options={fill=white, draw=Mrfuzzy, thick}] coordinates {(1.052,1.139) (1.116,1.130) (1.223,0.954) (1.008,1.044) (1.129,1.291) (1.004,1.164) (1.257,1.649) (1.007,1.069) (1.115,1.148) (1.253,1.074) (1.012,0.996) (1.119,1.114) (1.301,1.124) (1.007,1.047) (1.116,1.080)};
\addplot[only marks, mark=pentagon*, mark size=1.9pt, mark options={fill=Mspeccasctok, draw=Mspeccasctok}] coordinates {(1.052,0.875) (1.133,0.958) (1.060,1.002) (1.159,1.484) (1.008,1.050) (1.107,1.045) (1.050,1.004) (1.158,1.160) (0.992,0.971) (1.086,0.963) (1.040,1.080) (1.153,1.126)};
\addlegendentry{\texttt{spec\_casc\_tok}}
\addplot[forget plot, only marks, mark=pentagon*, mark size=1.9pt, mark options={fill=white, draw=Mspeccasctok, thick}] coordinates {(1.006,0.976) (1.021,0.992) (1.040,0.987) (1.019,0.964) (1.065,0.983) (1.000,1.059) (1.033,1.079) (1.010,1.008) (1.055,1.003) (1.010,1.018) (1.070,1.038) (1.011,1.051) (1.033,0.996)};
\addlegendimage{only marks, mark=*, mark size=1.9pt, mark options={fill=black, draw=black}}
\addlegendentry{GPT-OSS-20B}
\addlegendimage{only marks, mark=*, mark size=1.9pt, mark options={fill=white, draw=black, thick}}
\addlegendentry{Qwen3-8B}
\end{axis}
\end{tikzpicture}
\caption{Every tested setting of the primary pairs (151 runs; filled markers GPT-OSS-20B, hollow Qwen3-8B): length inflation against acceptance gain. A relaxed rule saves verifier rounds only below the dashed line $\lambda = (\laccept+1)/(\laccept^{*}+1)$ (Equation~\eqref{eq:netwin}). All twelve pairs, one panel per rule: Figure~\ref{fig:rules}.}
\label{fig:tradeoff}
\end{figure*}

\begin{figure*}[t]
\centering
\nolinenumbers
\begin{tikzpicture}
\begin{groupplot}[group style={group name=R, group size=5 by 1, horizontal sep=5pt, y descriptions at=edge left}, scale only axis, width=0.168\textwidth, height=0.15\textwidth, xmode=log, ymode=log, xmin=0.76, xmax=5.2, ymin=0.27, ymax=15, xtick={1,2,4}, ytick={0.5,1,2,4,8}, log ticks with fixed point, xminorticks=false, yminorticks=false, tick label style={font=\footnotesize}, label style={font=\footnotesize}, title style={font=\scriptsize, yshift=-4pt, text height=1.6ex, text depth=0.3ex}, grid=major, grid style={gray!20}, axis line style={gray!60}, every axis plot/.append style={mark size=1.2pt}]
\nextgroupplot[title={\texttt{mentored\_dec}}, ylabel={length inflation $\lambda$}]
\addplot[gray!60, dashed, thick, domain=0.76:5.2, samples=2, forget plot] {x};
\addplot[gray!60, dotted, forget plot] coordinates {(0.76,1) (5.2,1)};
\addplot[forget plot, only marks, mark=square*, mark options={fill=Dlm!55, draw=Dlm}] coordinates {(1.136,0.920) (1.159,0.941) (1.166,1.022) (1.180,0.906) (1.197,0.933) (1.220,0.921) (1.225,0.857) (1.248,0.833) (1.250,0.984) (1.250,1.052) (1.253,0.832) (1.266,0.909) (1.287,0.939) (1.298,1.024) (1.307,0.996) (1.325,1.255) (1.343,0.832) (1.349,1.007) (1.356,1.345) (1.358,0.912) (1.373,1.056) (1.389,0.348) (1.392,0.950) (1.401,1.182)};
\addplot[forget plot, only marks, mark=triangle*, mark options={fill=Dmid!55, draw=Dmid}] coordinates {(1.138,1.097) (1.155,0.953) (1.158,0.943) (1.166,1.020) (1.249,1.130) (1.278,1.012)};
\addplot[forget plot, only marks, mark=*, mark options={fill=Deagle!55, draw=Deagle}] coordinates {(1.013,1.033) (1.023,1.237) (1.025,1.027) (1.034,1.071) (1.055,1.029) (1.056,1.410) (1.059,1.070) (1.068,1.022) (1.090,1.036) (1.091,1.013) (1.099,1.031) (1.099,1.042) (1.108,1.282) (1.115,1.050) (1.135,1.630) (1.141,1.084) (1.147,1.431) (1.156,1.591) (1.173,1.705) (1.176,0.868) (1.194,2.128) (1.206,3.184) (1.216,1.109) (1.223,3.432) (1.231,1.479) (1.267,1.310) (1.285,1.087) (1.356,1.356) (1.382,2.084) (1.488,2.862)};
\addplot[forget plot, only marks, mark=*, mark size=1.5pt, mark options={fill=Deagle, draw=black, line width=0.5pt}] coordinates {(1.073,1.082) (1.074,1.062) (1.093,1.026) (1.094,1.012) (1.112,1.050) (1.156,1.039) (1.306,1.296) (1.327,1.163) (1.353,1.248) (1.359,1.695) (1.381,1.098) (1.408,1.595)};
\nextgroupplot[title={\texttt{cactus}}]
\addplot[gray!60, dashed, thick, domain=0.76:5.2, samples=2, forget plot] {x};
\addplot[gray!60, dotted, forget plot] coordinates {(0.76,1) (5.2,1)};
\addplot[forget plot, only marks, mark=square*, mark options={fill=Dlm!55, draw=Dlm}] coordinates {(1.182,0.893) (1.195,0.942) (1.205,0.933) (1.219,0.935) (1.253,0.851) (1.271,0.907) (1.282,1.444) (1.305,1.147) (1.314,0.910) (1.323,0.947) (1.326,0.808) (1.366,0.918) (1.379,0.756) (1.419,2.205) (1.446,1.487) (1.482,0.560) (1.492,0.913) (1.515,2.447) (1.593,2.453) (2.289,2.419) (2.307,1.880) (2.312,2.639) (2.400,2.426) (2.537,1.279)};
\addplot[forget plot, only marks, mark=triangle*, mark options={fill=Dmid!55, draw=Dmid}] coordinates {(1.140,1.022) (1.147,1.105) (1.166,0.935) (1.275,1.088) (1.293,0.973) (1.330,1.058)};
\addplot[forget plot, only marks, mark=*, mark options={fill=Deagle!55, draw=Deagle}] coordinates {(1.078,1.062) (1.099,1.042) (1.134,1.201) (1.143,1.096) (1.318,1.071) (1.357,1.460) (1.359,1.144) (1.442,13.235) (1.557,5.480) (1.592,1.971) (1.596,3.032) (1.632,5.601) (1.666,5.844) (1.675,4.294) (1.695,2.494) (1.715,1.315) (1.736,12.767) (1.842,9.372) (1.918,1.563) (1.939,2.554) (1.958,1.345) (1.971,8.131) (2.058,4.046) (2.502,4.408) (2.572,1.329) (3.080,2.596) (3.315,4.099) (4.155,2.916) (4.233,3.998) (4.381,13.218)};
\addplot[forget plot, only marks, mark=*, mark size=1.5pt, mark options={fill=Deagle, draw=black, line width=0.5pt}] coordinates {(1.133,1.145) (1.140,1.038) (1.506,1.609) (1.514,1.149) (1.519,1.791) (1.534,1.538) (1.544,1.779) (1.729,1.184) (1.790,1.156) (2.003,1.768) (2.121,1.257) (2.615,2.100)};
\nextgroupplot[title={\texttt{spec\_casc\_opt}}, xlabel={acceptance gain $(\laccept+1)/(\laccept^{*}+1)$}, legend style={at={(0.5,1.2)}, anchor=south, font=\scriptsize, draw=none, fill=none, column sep=6pt}, legend columns=4]
\addplot[gray!60, dashed, thick, domain=0.76:5.2, samples=2, forget plot] {x};
\addplot[gray!60, dotted, forget plot] coordinates {(0.76,1) (5.2,1)};
\addlegendimage{only marks, mark=*, mark options={fill=Deagle!55, draw=Deagle}}
\addlegendentry{EAGLE-3, EAGLE-1, P-EAGLE heads}
\addlegendimage{only marks, mark=triangle*, mark options={fill=Dmid!55, draw=Dmid}}
\addlegendentry{DSpark}
\addlegendimage{only marks, mark=square*, mark options={fill=Dlm!55, draw=Dlm}}
\addlegendentry{standalone drafters}
\addlegendimage{only marks, mark=*, mark size=1.5pt, mark options={fill=Deagle, draw=black, line width=0.5pt}}
\addlegendentry{primary pairs}
\addplot[forget plot, only marks, mark=square*, mark options={fill=Dlm!55, draw=Dlm}] coordinates {(1.079,0.946) (1.119,0.981) (1.128,0.948) (1.163,0.810) (1.174,1.070) (1.175,0.947) (1.217,0.803) (1.222,0.919) (1.241,0.871) (1.242,0.925) (1.245,0.915) (1.252,0.958) (1.277,0.906) (1.298,0.301) (1.314,1.070) (1.325,0.850) (1.344,3.166) (1.398,0.928) (1.416,1.091) (1.422,1.449) (1.456,1.397) (1.485,0.796) (1.523,1.579) (1.678,3.283)};
\addplot[forget plot, only marks, mark=triangle*, mark options={fill=Dmid!55, draw=Dmid}] coordinates {(1.148,1.042) (1.288,1.425) (1.306,1.104) (1.486,1.518) (1.495,1.155) (1.546,1.105)};
\addplot[forget plot, only marks, mark=*, mark options={fill=Deagle!55, draw=Deagle}] coordinates {(1.054,1.142) (1.065,1.024) (1.070,1.458) (1.084,1.042) (1.085,1.436) (1.094,1.077) (1.117,1.008) (1.123,1.015) (1.124,1.100) (1.150,1.280) (1.158,1.622) (1.183,1.112) (1.215,1.346) (1.254,1.313) (1.256,1.137) (1.282,1.900) (1.321,1.031) (1.322,1.582) (1.338,4.883) (1.387,1.231) (1.431,1.085) (1.556,2.585) (1.589,1.325) (1.592,2.869) (1.628,3.523) (1.641,1.883) (2.099,4.241) (2.907,2.702) (3.107,0.808) (4.848,8.260)};
\addplot[forget plot, only marks, mark=*, mark size=1.5pt, mark options={fill=Deagle, draw=black, line width=0.5pt}] coordinates {(1.224,1.269) (1.254,1.719) (1.268,1.328) (1.281,1.224) (1.291,1.734) (1.308,1.455) (1.329,1.788) (1.332,1.418) (1.342,1.091) (1.355,1.640) (1.373,2.484) (1.375,1.160)};
\nextgroupplot[title={\texttt{r\_fuzzy}}]
\addplot[gray!60, dashed, thick, domain=0.76:5.2, samples=2, forget plot] {x};
\addplot[gray!60, dotted, forget plot] coordinates {(0.76,1) (5.2,1)};
\addplot[forget plot, only marks, mark=square*, mark options={fill=Dlm!55, draw=Dlm}] coordinates {(0.985,0.656) (1.115,2.026) (1.198,0.879) (1.213,0.924) (1.216,0.940) (1.222,1.514) (1.261,0.787) (1.282,1.165) (1.288,0.849) (1.295,0.829) (1.298,0.924) (1.301,1.520) (1.316,1.199) (1.352,0.901) (1.362,1.063) (1.365,1.006) (1.422,1.028) (1.435,0.813) (1.446,1.366) (1.473,0.887) (1.474,0.546) (1.481,1.601) (1.492,1.021) (1.555,1.928)};
\addplot[forget plot, only marks, mark=triangle*, mark options={fill=Dmid!55, draw=Dmid}] coordinates {(1.154,1.068) (1.203,1.044) (1.223,1.244) (1.230,1.204) (1.254,1.060) (1.295,1.064)};
\addplot[forget plot, only marks, mark=*, mark options={fill=Deagle!55, draw=Deagle}] coordinates {(0.803,3.860) (0.808,3.729) (0.820,2.771) (0.867,1.677) (0.874,1.684) (0.955,1.058) (0.967,1.209) (0.968,1.193) (0.976,1.391) (0.982,1.188) (0.995,1.277) (1.000,0.989) (1.003,0.975) (1.008,1.117) (1.040,2.440) (1.084,3.593) (1.106,4.345) (1.130,1.501) (1.146,0.842) (1.156,1.138) (1.187,1.668) (1.188,1.283) (1.189,1.309) (1.201,1.222) (1.268,1.511) (1.270,1.514) (1.296,3.111) (1.387,1.103) (1.443,2.402) (1.568,1.575)};
\addplot[forget plot, only marks, mark=*, mark size=1.5pt, mark options={fill=Deagle, draw=black, line width=0.5pt}] coordinates {(1.116,1.080) (1.129,1.291) (1.223,0.954) (1.253,1.074) (1.257,1.649) (1.301,1.124) (1.352,1.638) (1.407,1.695) (1.425,1.653) (1.459,1.200) (1.466,2.132) (1.518,1.693)};
\nextgroupplot[title={\texttt{spec\_casc\_tok}}]
\addplot[gray!60, dashed, thick, domain=0.76:5.2, samples=2, forget plot] {x};
\addplot[gray!60, dotted, forget plot] coordinates {(0.76,1) (5.2,1)};
\addplot[forget plot, only marks, mark=square*, mark options={fill=Dlm!55, draw=Dlm}] coordinates {(1.034,0.947) (1.059,0.952) (1.065,0.935) (1.068,0.968) (1.094,0.827) (1.101,0.916) (1.102,0.975) (1.104,0.846) (1.107,0.943) (1.107,0.970) (1.121,0.936) (1.128,0.919) (1.133,0.586) (1.138,0.917) (1.141,0.934) (1.142,0.887) (1.146,0.788) (1.196,0.997) (1.209,0.929) (1.210,1.066) (1.224,0.862) (1.236,0.942) (1.302,1.037) (1.331,1.365)};
\addplot[forget plot, only marks, mark=triangle*, mark options={fill=Dmid!55, draw=Dmid}] coordinates {(1.074,1.047) (1.101,0.911) (1.141,0.942) (1.143,1.011) (1.155,0.964) (1.165,1.097)};
\addplot[forget plot, only marks, mark=*, mark options={fill=Deagle!55, draw=Deagle}] coordinates {(0.994,1.276) (1.000,1.013) (1.001,1.018) (1.003,1.060) (1.025,1.001) (1.028,1.094) (1.029,0.998) (1.031,0.825) (1.038,0.868) (1.042,0.991) (1.046,1.021) (1.049,0.946) (1.053,1.036) (1.057,1.071) (1.061,0.729) (1.063,0.802) (1.070,0.980) (1.073,1.109) (1.074,1.022) (1.082,1.029) (1.082,0.959) (1.092,0.861) (1.093,1.021) (1.104,0.830) (1.105,1.009) (1.115,0.842) (1.120,0.937) (1.142,0.865) (1.147,0.461) (1.426,0.396)};
\addplot[forget plot, only marks, mark=*, mark size=1.5pt, mark options={fill=Deagle, draw=black, line width=0.5pt}] coordinates {(1.033,0.996) (1.033,1.079) (1.040,0.987) (1.055,1.003) (1.065,0.983) (1.070,1.038) (1.086,0.963) (1.107,1.045) (1.133,0.958) (1.153,1.126) (1.158,1.160) (1.159,1.484)};
\end{groupplot}
\end{tikzpicture}
\caption{All 360 cells at each rule's most relaxed tested setting, one panel per rule (log scales; outlined: the primary pairs). A cell saves verifier rounds only below the dashed line $\lambda = (\laccept+1)/(\laccept^{*}+1)$ (Equation~\eqref{eq:netwin}). Above it: 21, 31, 31, 38, and 11 of each rule's 72 cells (mean gain 1.21, 1.74, 1.41, 1.22, 1.11), and 111 of 210 cells with an EAGLE-3, EAGLE-1, or P-EAGLE head, 3 of 30 with DSpark, and 18 of 120 with a standalone drafter.}
\label{fig:rules}
\end{figure*}
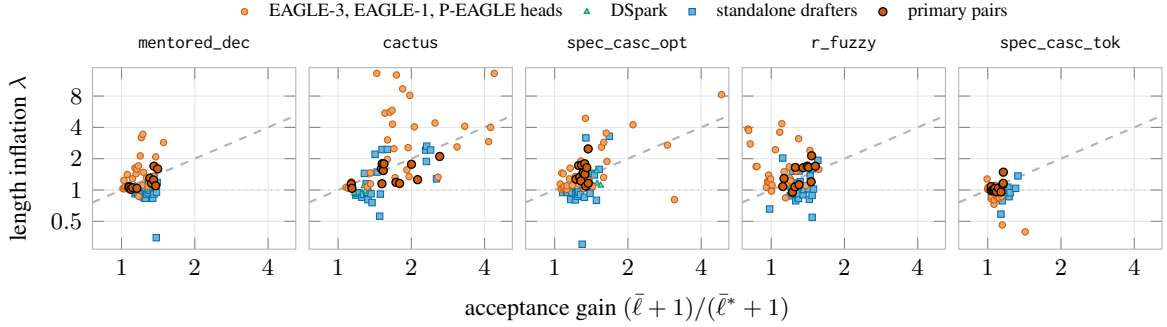

The extra acceptance buys few rounds (Tables~\ref{tab:rounds} and~\ref{tab:pairs}, Figures~\ref{fig:tradeoff} and~\ref{fig:rules}): over the 360 cells $R/R^{*}$ has median $0.93$ and mean $1.10$, and 114 cells need \emph{more} rounds than lossless (67 significantly). The round ratio follows length, not acceptance (Spearman $\rho$ $0.78$ with $\lambda$, $-0.12$ with the acceptance gain), and the length-corrected model of Equation~\eqref{eq:speedup-relax}, which at fixed $\Ndraft$ reduces to Equation~\eqref{eq:netwin}, predicts it from the measured length within $2\%$ at the median.

Time is worse than rounds on the primary pairs (median ratio $1.01$ against $0.96$; eleven cells save rounds but not time) because on their server a round's cost grows with the tokens it commits; on a second server they agree (Appendix~\ref{app:seeds}), so rounds are the hardware-independent measure.

\begin{table*}[t]
\centering\scriptsize
\renewcommand{\arraystretch}{0.9}
\setlength{\tabcolsep}{3pt}
\begin{tabular}{@{}l ccccc ccccc@{}}
\toprule
Dataset & ment. & cactus & c-opt & r-fuzzy & c-tok & ment. & cactus & c-opt & r-fuzzy & c-tok \\
\midrule
 & \multicolumn{5}{c}{GPT-OSS-20B + EAGLE-3 (nebius)} & \multicolumn{5}{c}{Qwen3-8B + EAGLE-3 (RedHatAI)} \\
\cmidrule(lr){2-6}\cmidrule(lr){7-11}
GSM8K & \underline{0.83} & 0.99 & 1.07 & 1.14 & \underline{0.83} & \underline{0.93} & \underline{0.90} & 1.05 & \underline{0.76} & \underline{0.95} \\
AIME24 & 1.18 & 1.07 & \textbf{1.70} & \textbf{1.37} & 1.22 & 0.94 & \underline{0.55} & 1.06 & 1.16 & 0.93 \\
HumanEval & 0.97 & \textbf{1.12} & \textbf{1.29} & \textbf{1.16} & 0.93 & 0.99 & 0.96 & \textbf{1.36} & \textbf{1.33} & 1.04 \\
LiveCodeBench & \underline{0.89} & 0.94 & \textbf{1.16} & 1.11 & 0.98 & \underline{0.93} & \underline{0.72} & 1.03 & \underline{0.87} & \underline{0.94} \\
MT-Bench & \underline{0.75} & \underline{0.62} & \underline{0.76} & \underline{0.76} & \underline{0.88} & \underline{0.89} & \underline{0.64} & 0.94 & \underline{0.85} & 0.96 \\
LongBench-v2 & 1.11 & 0.87 & \textbf{1.31} & 1.10 & 0.97 & 0.98 & \underline{0.80} & \underline{0.83} & 0.96 & 0.96 \\
\bottomrule
\end{tabular}
\caption{Verifier rounds relative to lossless ($R/R^{*}$) on the primary pairs at each method's most relaxed tested setting (ment., c-: mentored, casc-). Bold: the 95\% interval for $R/R^{*}$ lies entirely above $1$ (more rounds than lossless); underline: entirely below $1$. All twelve pairs: Table~\ref{tab:rounds-all}.}
\label{tab:rounds}
\end{table*}

\subsection{Accuracy retention}
\label{sec:accuracy}

Relaxation costs accuracy in 198 of the 300 graded cells (lossless references: Tables~\ref{tab:strict} and~\ref{tab:strict-all}; MT-Bench: Appendix~\ref{app:results}). The three aggressive rules lose in 143 of their 180 graded cells, by $0.19$ on average for \texttt{cactus}, $0.14$ for \texttt{spec\_casc\_opt}, and $0.21$ for \texttt{r\_fuzzy}; \texttt{mentored\_dec} loses in 33 of 60 cells (mean $0.03$) and \texttt{spec\_casc\_tok} in 22 of 60 (mean $0.005$).

The losses follow the inflation: 132 of the 144 graded cells whose $\lambda$ interval lies above $1$ lose accuracy (mean loss $0.21$), against 66 of the other 156 (mean $0.03$). By task they are largest on code generation, with median losses of $0.09$ on HumanEval and $0.11$ on LiveCodeBench against $0.03$ to $0.04$ on the other graded datasets.

\subsection{Analysis}
\label{sec:analysis}

\finding{Settings that need fewer verifier rounds lose accuracy.} With an EAGLE head, the three aggressive rules need fewer rounds than lossless in 40 of their 105 graded cells, and 35 of these lose accuracy, most where the cut is deepest: \texttt{cactus} cuts rounds to $0.44\times$ on R1-Distill-Llama-8B/LiveCodeBench and to $0.55\times$ on Qwen3-8B/AIME24, while accuracy falls from $0.58$ to $0.02$ and from $0.70$ to $0.23$. Of the 86 graded cells that save rounds and keep accuracy, 68 use \texttt{spec\_casc\_tok}, DSpark, or a standalone Qwen3 drafter, the choices closest to the target (below). On the primary pairs, tuning for accuracy saves a few percent of the rounds at best: only 29 of the 50 graded cells have a tested setting within two points of lossless, and the best of these have a median round ratio of $0.96$ (Table~\ref{tab:best}).

\finding{Keeping accuracy means staying close to the target.} \texttt{spec\_casc\_tok}, the only rule that can reject what lossless verification accepts (Section~\ref{sec:trainingfree}), is also the only one that keeps accuracy on average: over its 60 graded cells it loses $0.005$, against $0.03$ to $0.21$ for the other rules, and its $\lambda$ interval lies above $1$ in 5 of its 72 cells, against 28 to 51. Its acceptance gain is the smallest ($1.11\times$), yet because its completions do not grow (mean $\lambda$ $0.96$) it needs $0.86\times$ the rounds of lossless on average.

\begin{table*}[!t]
\centering\scriptsize
\renewcommand{\arraystretch}{0.9}
\setlength{\tabcolsep}{2.5pt}
\begin{tabular}{@{}l rr rrrrrr rrrrrr r@{}}
\toprule
 & & & \multicolumn{6}{c}{mean $\lambda$} & \multicolumn{6}{c}{mean $R/R^{*}$} & \\
\cmidrule(lr){4-9}\cmidrule(lr){10-15}
Target, drafter & $\laccept^{*}$ & gain & ment. & cactus & c-opt & r-fuzzy & c-tok & all & ment. & cactus & c-opt & r-fuzzy & c-tok & all & acc.\ loss \\
\midrule
\multicolumn{16}{@{}l}{GPT-OSS-20B} \\
\quad EAGLE-3 (nebius) & 2.28 & 1.38 & 1.35 & \textbf{1.61} & 1.70 & \textbf{1.67} & 1.12 & 1.49 & 0.96 & 0.93 & 1.22 & 1.10 & 0.97 & 1.04 & 22/25 \\
\quad EAGLE-3 (RedHatAI) & 1.49 & 1.35 & 1.31 & 2.32 & 1.36 & 1.84 & 1.07 & 1.58 & 1.14 & 1.01 & 1.16 & 1.52 & 1.05 & 1.18 & 17/25 \\
\addlinespace[1pt]
\multicolumn{16}{@{}l}{Qwen3-8B} \\
\quad EAGLE-3 (RedHatAI) & 1.20 & 1.27 & 1.05 & 1.31 & \textbf{1.35} & 1.20 & 1.01 & 1.18 & 0.94 & 0.76 & 1.04 & 0.99 & 0.96 & 0.94 & 18/25 \\
\quad P-EAGLE (RedHatAI) & 1.88 & 1.12 & 1.03 & 1.10 & 1.07 & \textbf{1.35} & 1.00 & 1.11 & 0.95 & 0.93 & 0.97 & 1.15 & 0.96 & 0.99 & 24/25 \\
\quad DSpark (DeepSeek) & 2.35 & 1.23 & 1.03 & 1.03 & \textbf{1.23} & 1.11 & 1.00 & 1.08 & 0.85 & 0.83 & 0.87 & 0.90 & 0.87 & 0.86 & 18/25 \\
\quad Qwen3-0.6B & 2.92 & 1.24 & 0.92 & 0.90 & 0.92 & 0.97 & 0.94 & 0.93 & \underline{0.75} & \underline{0.67} & \underline{0.78} & \underline{0.67} & \underline{0.87} & 0.75 & 6/25 \\
\quad Qwen3-1.7B & 3.60 & 1.25 & \underline{0.89} & \underline{0.89} & 0.91 & 0.89 & 0.91 & 0.90 & \underline{0.69} & \underline{0.69} & \underline{0.69} & \underline{0.69} & \underline{0.78} & 0.71 & 4/25 \\
\addlinespace[1pt]
\multicolumn{16}{@{}l}{Llama-3.1-8B-Instruct} \\
\quad EAGLE-3 (yuhuili) & 1.67 & 1.57 & 1.94 & \textbf{8.85} & 2.70 & 2.15 & 0.84 & 3.30 & 1.51 & 3.96 & 1.01 & 3.00 & 0.70 & 2.03 & 18/25 \\
\quad EAGLE-1 (yuhuili) & 1.29 & 1.42 & 1.95 & \textbf{6.23} & \textbf{3.35} & 2.58 & 0.78 & 2.98 & 1.60 & \textbf{3.16} & 1.57 & 2.60 & 0.70 & 1.93 & 19/25 \\
\quad Llama-3.2-1B & 3.57 & 1.32 & 0.94 & 1.59 & 0.83 & 1.30 & 0.85 & 1.10 & 0.70 & 1.08 & \underline{0.66} & 0.90 & 0.74 & 0.82 & 16/25 \\
\addlinespace[1pt]
\multicolumn{16}{@{}l}{R1-Distill-Llama-8B} \\
\quad EAGLE-3 (yuhuili) & 1.42 & 1.41 & 1.04 & \textbf{2.11} & 1.19 & 1.19 & 0.98 & 1.30 & 0.92 & 0.80 & 0.84 & 1.05 & 0.93 & 0.91 & 16/25 \\
\quad Llama-3.2-1B & 1.84 & 1.48 & 1.07 & \textbf{2.14} & \textbf{1.99} & 1.24 & 1.04 & 1.50 & 0.80 & 0.95 & 1.34 & 1.10 & 0.83 & 1.01 & 20/25 \\
\midrule
All twelve pairs & --- & 1.34 & 1.21 & 2.51 & 1.55 & 1.46 & 0.96 & 1.54 & 0.98 & 1.31 & 1.01 & 1.31 & 0.86 & 1.10 & 198/300 \\
\bottomrule
\end{tabular}
\caption{Twelve target-drafter pairs at each rule's most relaxed tested setting: for each rule, the mean of $\lambda$ and of $R/R^{*}$ over the pair's datasets; all: over all the pair's cells. Bold (underline): the 95\% interval lies above (below) $1$ on every dataset. $\laccept^{*}$: mean lossless accepted length; gain: mean $(\laccept+1)/(\laccept^{*}+1)$; acc.\ loss: graded cells two or more points below lossless. Every cell: Tables~\ref{tab:length-all} and~\ref{tab:rounds-all}. ment., c-: mentored, casc-.}
\label{tab:pairs}
\end{table*}

\finding{The inflation is a property of the drafter, not of the target.} The same target inflates very differently with each drafter (Table~\ref{tab:pairs}, Figure~\ref{fig:rules}): Llama-3.1-8B-Instruct by $3.30\times$ on average with its EAGLE-3 head and $1.10\times$ with Llama-3.2-1B, Qwen3-8B by $1.18\times$ with its EAGLE-3 head and $0.90\times$ with Qwen3-1.7B. Every EAGLE head inflates ($\lambda$ interval above $1$ in 132 of 210 cells, below $1$ in 8), and Llama-3.1-8B-Instruct with its EAGLE heads is the worst case: its lossless answers run 280 to 4\,840 tokens, and \texttt{cactus} multiplies them by $4.0$ to $13.2$. Drafters that sample close to the target inflate little: DSpark has median $\lambda$ $1.06$ and a round interval below $1$ in 22 of its 30 cells, and the standalone Qwen3 drafters lengthen completions significantly in none of their 60 cells and cut rounds by $10\%$ to $46\%$, losing accuracy in 10 of 50 graded cells against 18 and 24 of 25 for DSpark and P-EAGLE. A standalone drafter from another model's distribution inflates again: Llama-3.2-1B on R1-Distill-Llama-8B, up to $3.28\times$ on LongBench-v2.

\subsection{Why completions get longer}
\label{sec:why}

For all twelve pairs the run logs record whether a run hit its budget; for the primary pairs they also record the split between thinking trace and answer and, for 15 traced GPT-OSS-20B cases per dataset and arm, every drafted token's target probability and both rules' verdicts. Four measurements follow, at the loosest settings.

\finding{The extra text is thinking, not answer.} On the primary pairs, GPT-OSS-20B's analysis channel receives at least $92\%$ of the extra characters in 25 of its 30 cells. Qwen3-8B puts more than $90\%$ of the extra characters inside its \texttt{<think>} blocks at the median, with one exception: \texttt{cactus} on AIME24 inflates the answer ($9.4\times$) because 11 of its 30 outputs reopen a closed think block. Llama-3.1-8B-Instruct has no thinking trace, so its inflation, the largest measured, lengthens the answer itself.

\finding{A broad shift with a heavy tail.} On the primary pairs the per-case ratio $L_{\mathrm{relax}}/L_{\mathrm{strict}}$ has a median above $1$ in 47 of the 60 cells and a lower quartile below $1$ in 34. Its 90th percentile has median $2.2$ over the cells, and the tenth of cases with the largest increase holds $59\%$ of the net extra tokens at the median. This is not exact looping: on GPT-OSS-20B at most $2.9\%$ of a cell's tokens lie inside a repeated 50-gram (lossless: at most $1.9\%$). Qwen3-8B does loop under \texttt{spec\_casc\_opt}: $15\%$ of its AIME24 tokens and $13\%$ of its LongBench-v2 tokens (lossless $0.4\%$ and $1.3\%$), with a mean longest repeat of 2\,728 tokens on AIME24.

\finding{Cut-offs understate inflation.} Relaxation raises the share of runs that end at the generation budget in 208 of the 360 cells (median $11\%$ against $4\%$ under lossless, where Qwen3-8B already reaches $25\%$ on GSM8K; per-cell shares in the released tables). On the primary pairs, keeping only the cases that finished under both rules moves $\lambda$ by at most $0.1$ in 45 of the 60 cells, so the inflation is not an artifact of the cap.

\finding{Relaxation admits tokens the target rates as unlikely.} We split the committed draft tokens of the traced runs into those both rules accept and those only the relaxed rule accepts (Table~\ref{tab:admit}). The relaxed-only tokens are $7\%$ to $24\%$ of the committed draft tokens. For the four rules other than \texttt{spec\_casc\_tok} they have a median target probability of $0.01$ to $0.09$ (shared tokens: $0.78$ to $0.93$) and a mean target rank of 37 to 234. \texttt{spec\_casc\_tok} admits the fewest extra tokens ($7\%$), and they are the target's first or second choice (mean rank $0.9$, median probability $0.29$). Lossless verification accepts any token with $q(x) \le p(x)$, so a relaxed rule can only add tokens with $q(x) > p(x)$: the overshoot tokens that \citet{wang2026revisiting} tie to quality loss. In our traces they also carry the extra length. A lossless control agrees: raising the sampling temperature from $1.0$ to $1.2$, which also commits lower-probability tokens, lengthens GPT-OSS-20B completions by $1.27\times$ on GSM8K and $1.07\times$ on LiveCodeBench, and $T = 1.5$ inflates GSM8K by $3.9\times$ with accuracy $0.31$ (Appendix~\ref{app:seeds}). The logs show two things: relaxation commits tokens the target would rarely sample, and the thinking trace then gets longer. They do not show why an unlikely token makes a reasoning model think more rather than less. % TODO(Haochen): a per-token account of what follows a relaxed-only token (does the model correct it, restart the step, or continue?) would close this gap; add it here if the traces support it.

\section{What the Extra Length Buys}
\label{sec:buys}

Reasoning models often answer better when they think longer \citep{muennighoff2025s1}, so the length that relaxation adds could be useful thinking. Per-token guards that force lossless verification do not shorten completions beyond their tuning cases (Appendix~\ref{app:buys}), so we act on whole runs.

\finding{Closing the reasoning trace recovers answers.} We rerun every relaxed run that passes a token threshold near the generation budget without closing its reasoning trace, and at the threshold force the token that closes the trace, the budget forcing of \citet{muennighoff2025s1}; runs that close their trace earlier are unchanged (Appendix~\ref{app:buys}). Over 1\,132 such runs (five rules, both primary pairs, five graded datasets; Table~\ref{tab:force}), correct answers rise from 126 to 439: 335 runs that failed now answer correctly and 22 that succeeded now fail, 302 and 7 of them on Qwen3-8B.

\begin{table}[t]
\centering\scriptsize
\setlength{\tabcolsep}{3pt}
\begin{tabular}{@{}l rr@{\,$\to$\,}lrr rr@{\,$\to$\,}lrr@{}}
\toprule
 & \multicolumn{5}{c}{Qwen3-8B} & \multicolumn{5}{c}{GPT-OSS-20B} \\
\cmidrule(lr){2-6}\cmidrule(lr){7-11}
Dataset & forced & \multicolumn{2}{c}{correct} & $+$ & $-$ & forced & \multicolumn{2}{c}{correct} & $+$ & $-$ \\
\midrule
GSM8K & 307 & 56 & 225 & 171 & 2 & 32 & 4 & 10 & 6 & 0 \\
AIME24 & 60 & 2 & 6 & 4 & 0 & 36 & 3 & 7 & 7 & 3 \\
HumanEval & 205 & 16 & 52 & 37 & 1 & 9 & 2 & 0 & 0 & 2 \\
LiveCodeBench & 226 & 11 & 81 & 72 & 2 & 62 & 7 & 16 & 13 & 4 \\
LongBench-v2 & 157 & 17 & 33 & 18 & 2 & 38 & 8 & 9 & 7 & 6 \\
\midrule
Total & 955 & 102 & 397 & 302 & 7 & 177 & 24 & 42 & 33 & 15 \\
\bottomrule
\end{tabular}
\caption{Forcing relaxed runs to close their reasoning trace at a threshold near the generation budget, summed over the five rules at their loosest tested settings. \emph{forced}: runs still reasoning at the threshold; \emph{correct}: how many of them are correct without~$\to$ with forcing; $+$: forced runs that go from wrong to correct; $-$: from correct to wrong. Per rule and thresholds: Table~\ref{tab:force-all}.}
\label{tab:force}
\end{table}

\finding{The extra tokens follow the answer on grade-school math and precede it on competition math.} For cases both arms answer correctly, we split the extra tokens at the first point where the output writes the reference answer (Table~\ref{tab:timing}). On GSM8K, $50\%$ (GPT-OSS-20B) and $82\%$ (Qwen3-8B) of the extra tokens come after it, and most runs cut off at the budget already contain the answer. On AIME24, relaxed runs reach the answer later (median token 3\,865 against 1\,721 for GPT-OSS-20B), at least $59\%$ of the extra tokens precede it, and few cut-off runs contain it.

\finding{The extra length is waste.} On problems within the model's reach, relaxation makes the model continue past an answer it already holds, as lossless reasoning models already do \citep{chen2025overthinking}; on hard problems, it lengthens the route to the answer. In neither case do the extra tokens buy accuracy.

\section{Conclusion}
\label{sec:conclusion}

We evaluated five training-free relaxed verification rules on six benchmarks and twelve target-drafter pairs, measuring completion length, verifier rounds, time, and accuracy rather than acceptance alone. Relaxation does raise the number of draft tokens accepted per round, but the model spends that gain on longer outputs: across the 360 cells the mean length inflation ($1.54\times$) exceeds the mean acceptance gain ($1.34\times$). Where a rule does decode faster, accuracy drops; the one rule that keeps accuracy close to lossless, \texttt{spec\_casc\_tok}, is also the one that relaxes least.

This paper identifies length as what decides the speed of relaxed verification: once our measured length is supplied, the length-corrected model of \citet{xia2026practical} predicts our verifier rounds within about $2\%$, and the clock departs from it only through the cost of a round, which the model does not contain. Two conclusions follow. Acceptance gain is not a measure of speed for a relaxed rule, because relaxation changes what the model generates and not only how fast it is verified; completion length and verifier rounds have to be reported next to it. And the promising direction is not accepting more of the drafter's tokens but choosing which ones to admit, since the only rule that worked trusted the target more.

Section 4 also yields a simple mitigation: closing the reasoning trace before the generation budget runs out recovers the answers that inflation cuts off, raising correct answers from 126 to 439 over the 1,132 runs it touches, under all five rules. It does not recover the speed, since these runs still spend most of their budget; closing the trace earlier, where the answer is already written, is the natural next step.

\section*{Limitations}

\paragraph{Matching and settings.}
No $\laccept$ is reachable by every method, so the matching of Section~\ref{sec:protocol} is best-effort and the tables compare methods at different $\laccept$, each at its most relaxed setting. Each method is tested at four published or reference $\alpha$ values; other values could change its profile, but the conclusions also hold at the gentlest and the best accuracy-preserving settings (Section~\ref{sec:length}; Tables~\ref{tab:gentle} and~\ref{tab:best}).

\paragraph{Budget censoring.}
Relaxed runs hit the per-dataset token budget more often, so lengths are right-censored and the inflation is, if anything, understated; part of the accuracy loss is unfinished runs. The GSM8K budget (2\,048 tokens) is tight for Qwen3-8B: $25\%$ of its lossless runs hit it (Table~\ref{tab:strict}).

\paragraph{Sampling and timing.}
Each case runs once per arm (seeds repeated on the primary pairs only, Section~\ref{sec:length}), and paired bootstrap intervals over cases are wide on AIME24 (30 cases) and MT-Bench (80), so we interpret only effects that recur across datasets or targets. Times are client-side at batch size 1 and analyzed on the primary pairs only; their ratio to lossless depends on the server's cost per committed token (Section~\ref{sec:rounds}) and is noisy because each benchmark's first cases ran on a fresh server process and the rest on a reused one, while round counts depend on neither. MT-Bench quality is one judge's first-turn score, not an accuracy.

\paragraph{Scope.}
We use one serving stack (vLLM 0.26.0) and, for the main results, one sampler (temperature $1.0$, top-$p$ $1.0$); on Qwen3-8B the gentle rules' inflation depends on the sampler (Section~\ref{sec:length}). Every relaxed arm drafts $\Ndraft = 6$ tokens; draft length is swept only under lossless verification (Appendix~\ref{app:seeds}). The rules are our reimplementations in vLLM's rejection sampler, checked at the strict limit but not against the authors' code; the ensemble rule, speculative contrastive decoding (Table~\ref{tab:taxonomy}) and typical acceptance \citep{cai2024medusa} are not swept. Lossless decoding at a higher temperature also inflates completions (Section~\ref{sec:why}), so the inflation is not specific to relaxed acceptance, and we do not explain why an unlikely token makes a reasoning model think longer.

% acl.sty already selects acl_natbib
\bibliography{references}

\appendix
% Appendix floats: pack the tables more tightly than in the body.
\renewcommand{\dbltopfraction}{0.9}
\renewcommand{\topfraction}{0.9}
\renewcommand{\textfraction}{0.05}
\renewcommand{\floatpagefraction}{0.75}
\renewcommand{\dblfloatpagefraction}{0.75}
\setcounter{dbltopnumber}{3}
\setcounter{topnumber}{4}
\setcounter{totalnumber}{6}
% Appendix tables are numbered per appendix: Table A1, B1, ...
\counterwithin{table}{section}
\renewcommand{\thetable}{\thesection\arabic{table}}

\section{Method Details}
\label{app:methods}

Algorithm~\ref{alg:specdec} gives one round of speculative decoding with the three relaxed target distributions in place, and Table~\ref{tab:taxonomy} states each training-free rule as a choice of them. The paragraphs below give our implementation of the five swept rules; Table~\ref{tab:evidence} collects what the source papers measure.

\begin{algorithm*}[t]
\caption{One round of speculative decoding with relaxed target distributions. Lossless (strict) speculative decoding is the special case $\pirej = \pires = \pibonus = p$; a lossy (relaxed) method is a choice of the three $\pi$ (Table~\ref{tab:taxonomy}).}
\label{alg:specdec}
\small
\begin{algorithmic}
\Require drafter $q$, target $p$, context $x_{<t}$, draft length $\Ndraft$, relaxation parameter $\alpha$, relaxed target distributions $\pirej, \pires, \pibonus$
\LineComment{\textbf{DRAFT}: sample $\Ndraft$ tokens autoregressively from $q$}
\For{$i = 0$ to $\Ndraft - 1$}
    \State $q_{t+i}(\cdot) = q(\cdot \mid x_{<t+i})$; \quad $x_{t+i} \sim q_{t+i}(\cdot)$
\EndFor
\LineComment{\textbf{VERIFY}: score all $\Ndraft + 1$ positions with one forward pass of $p$}
\State $p_{t+i}(\cdot) = p(\cdot \mid x_{<t+i})$ for $i \in \{0,\ldots,\Ndraft\}$
\LineComment{Construct the relaxed targets (strict decoding: $\pi = p$)}
\State $\pi^{\mathrm{rej}}_{t+i} = \pirej[q_{t+i}, p_{t+i}, x_{t+i}, \alpha]$, \; $\pi^{\mathrm{res}}_{t+i} = \pires[q_{t+i}, p_{t+i}, x_{t+i}, \alpha]$ for $i \in \{0,\ldots,\Ndraft-1\}$
\LineComment{Accept or reject each draft token; $i_{\mathrm{end}}$ is the first rejection}
\State $a_i \sim \Ber\!\left(\min\!\left\{1, \pi^{\mathrm{rej}}_{t+i}(x_{t+i}) / q_{t+i}(x_{t+i})\right\}\right)$ for $i \in \{0,\ldots,\Ndraft-1\}$; \; $a_{\Ndraft} = 0$
\State $i_{\mathrm{end}} = \min\{\, i \in \{0,\ldots,\Ndraft\} : a_i = 0 \,\}$
\If{$i_{\mathrm{end}} < \Ndraft$}
    \LineComment{Rejection: resample the rejected position from the residual}
    \State $x_{t+i_{\mathrm{end}}} \sim \normop\!\left(\max\{0,\, \pi^{\mathrm{res}}_{t+i_{\mathrm{end}}}(\cdot) - q_{t+i_{\mathrm{end}}}(\cdot)\}\right)$
\Else
    \LineComment{All accepted: bonus token (an extra drafter step gives $q_{t+\Ndraft}$ if $\pibonus$ needs it)}
    \State $x_{t+\Ndraft} \sim \pibonus[q_{t+\Ndraft}, p_{t+\Ndraft}, \alpha](\cdot)$
\EndIf
\State \Return $x_t, \ldots, x_{t+i_{\mathrm{end}}}$ \quad (between $1$ and $\Ndraft + 1$ tokens)
\end{algorithmic}
\end{algorithm*}

\begin{table*}[t]
\centering
\scriptsize
\setlength{\tabcolsep}{3pt}
\renewcommand{\arraystretch}{1.15}
\begin{tabular}{@{}>{\raggedright\arraybackslash}p{2.7cm} >{\raggedright\arraybackslash}p{2.1cm} >{\raggedright\arraybackslash}p{6.0cm} c c >{\raggedright\arraybackslash}p{1.5cm}@{}}
\toprule
Method (arm) & $\alpha$ controls & $\pirej[q,p,x,\alpha](v)$ & $\pires$ & $\pibonus$ & $\alpha$ swept \\
\midrule
CACTUS (\texttt{cactus}) \newline \citep{hao2026cactus}
  & reverse-KL budget $\KL(\pi \,\|\, p) \le \alpha$
  & $\gamma_x$ if $v = x$, else $\frac{1-\gamma_x}{1-p(x)}\,p(v)$; \newline $\gamma_x = \min\{p(x) + \sqrt{2\alpha\, p(x)(1-p(x))},\, 1\}$
  & $\pirej$ & $p$ & 0.03, 0.08, 0.18, 0.35 \\
Mentored decoding (\texttt{mentored\_dec}) \newline \citep{tranthien2023mentored}
  & forward-KL budget $\KL(p \,\|\, \pi) \le \alpha$
  & $\min\{p(v)/(1-\alpha),\, 1\}$
  & $p$ & $p$ & 0.15, 0.35, 0.55, 0.75 \\
Fuzzy, reducible variant (\texttt{r\_fuzzy}) \newline \citep{holsman2025fuzzy}
  & divergence threshold
  & $q(v)$ if $\Div(p, q) < \alpha$, else $p(v)$
  & $\pirej$ & $p$ & 0.03, 0.08, 0.15, 0.25 \\
Speculative cascade, optimal rule (\texttt{spec\_casc\_opt}) \newline \citep{narasimhan2025cascades}
  & deferral cost
  & $q(v)$ if $\max_u q(u) \ge \max_u p(u) - \alpha\,\TV(p, q)$, else $p(v)$
  & $\pirej$ & $p$ & $-0.3$, $-0.1$, $-0.02$, $0.05$ \\
Speculative cascade, token rule (\texttt{spec\_casc\_tok}) \newline \citep{narasimhan2025cascades}
  & trusted-set width
  & $q(v) + \eta_\alpha\, p(v)$ if $v \in A_\alpha$, else $\eta_\alpha\, p(v)$; \newline $A_\alpha = \{u : p(u) \ge (1-\alpha)\max_w p(w)\}$, $\eta_\alpha = 1 - \sum_{u \in A_\alpha} q(u)$
  & $\pirej$ & $p$ & 0.15, 0.35, 0.55, 0.8 \\
Ensemble (\texttt{ens}) \newline \citep{wang2026diversed}
  & mixture weight
  & $\alpha\, q(v) + (1-\alpha)\, p(v)$
  & $\pirej$ & $p$ & not swept \\
Speculative contrastive decoding (\texttt{spec\_cont\_dec}) \newline \citep{yuan2024scd}
  & contrast strength
  & $\softmax[z_\pi](v)$, \newline $z_\pi(v) = z_p(v) + \alpha\,[z_p(v) - z_q(v)]$ on the top-$p$/top-$k$ mask of $p$, $-\infty$ outside
  & $\pirej$ & $\pirej$ & not swept \\
\bottomrule
\end{tabular}
\caption{Training-free relaxed verification rules as instances of Algorithm~\ref{alg:specdec}. Strict speculative decoding is $\pi = p$ throughout. Every rule except the two cascade rules recovers strict verification at $\alpha = 0$; \texttt{spec\_casc\_opt} recovers it for $\alpha < -1$ (since $\max_u q(u) - \max_u p(u) \le \TV(p, q)$) and \texttt{spec\_casc\_tok} for $\alpha < 0$. On the second sampler path (Section~\ref{sec:setup}), \texttt{cactus} and \texttt{spec\_casc\_tok} use $\pires = p$. Last column: the four $\alpha$ values of the pilot (Section~\ref{sec:protocol}), used for every pair.}
\label{tab:taxonomy}
\end{table*}

\paragraph{CACTUS.}
\citet{hao2026cactus} choose, for the drafted token $x$, the relaxed target that maximizes the acceptance probability under a reverse-KL budget $\KL(\pi \,\|\, p) \le \alpha$; a Taylor expansion of the constraint gives the closed form of Table~\ref{tab:taxonomy}, which boosts $x$ as a function of $p(x)$ and shrinks every other token proportionally. The paper does not specify $\pibonus$; we use $p$, as the released code does. A draft token outside the target's top-$p$/top-$k$ support is rejected regardless of $\alpha$.

\paragraph{Mentored decoding.}
\citet{tranthien2023mentored} maximizes the expected acceptance probability subject to a forward-KL budget $\KL(p \,\|\, \pi) \le \alpha$. We implement the simplified form used by \citet[App.~F.6]{narasimhan2025cascades} and \citet{xia2026practical}: every token receives the same multiplicative boost, capped at one, and the residual after a rejection is the unrelaxed $p$.

\paragraph{Reducible fuzzy speculative decoding.}
\citet{holsman2025fuzzy} trust the drafter outright when the Jensen--Shannon divergence between $p$ and $q$, computed over the full vocabulary, is below $\alpha$. In their main variant (FSD) a failed test samples from $p$ directly; we implement the reducible variant rFSD (their Sec.~4.3), which falls back to the standard acceptance coin and so reduces to strict verification at $\alpha = 0$. \citet[App.~C]{xia2026practical} prove that rFSD samples from the same per-token distribution as FSD with weakly higher acceptance.

\paragraph{Speculative cascades.}
For \texttt{spec\_casc\_opt} we implement the plug-in deferral rule of \citet{narasimhan2025cascades}, which trusts the drafter when $\max_u q(u) \ge \max_u p(u) - \alpha\,\TV(p, q)$; at $\alpha = 0$ it still trusts the drafter wherever $\max_u q(u) \ge \max_u p(u)$, so the pilot includes negative $\alpha$. For \texttt{spec\_casc\_tok} we implement the token-specific rule (their Sec.~4.4) in the form of \citet[App.~B]{xia2026practical}: draft tokens in the trusted set $A_\alpha$ are accepted with probability one, and the others are judged against $\eta_\alpha\, p(v) \le p(v)$, which is stricter than lossless verification. Since the argmax of $p$ is always trusted, this rule is not strict even at $\alpha = 0$.

\subsection{What the source papers measure}
\label{app:evidence}

Table~\ref{tab:evidence} summarizes the results behind each method. The reported quantities are acceptance or accepted length, tokens per second or relative latency, and task quality, on short-form benchmarks: maximum output lengths of a few hundred tokens where they are stated at all, and timing budgets that fix the number of generated tokens in one case. Only \citet{xia2026practical} treat response length explicitly, as the correction term of Equation~\eqref{eq:speedup-relax} and as a finding that relaxation with weak drafters can lengthen responses enough to reduce speedup. CACTUS flags the same effect as a caveat and adds a length-sensitive proxy, but does not measure length.

\begin{table*}[t]
\centering
\footnotesize
\setlength{\tabcolsep}{4pt}
\renewcommand{\arraystretch}{1.2}
\begin{tabular}{@{}>{\raggedright\arraybackslash}p{3.0cm} >{\raggedright\arraybackslash}p{4.3cm} >{\raggedright\arraybackslash}p{3.9cm} c c c@{}}
\toprule
Method & Models and tasks & Speed metric & Length & Rounds & Long-form \\
\midrule
CACTUS \citep{hao2026cactus} & Qwen3, Gemma, R1-Distill, Llama pairs; GSM8K, IFEval, GPQA & $\mathrm{AL}_m$, rejected-token count, vLLM wall-clock & (\checkmark) & (\checkmark) & (\checkmark) \\
Mentored decoding \citep{tranthien2023mentored} & T5-large/small; WMT15 En--Fr & tokens per second & $\times$ & $\times$ & $\times$ \\
Fuzzy \citep{holsman2025fuzzy} & Llama-3.1, Gemma-2, Qwen2.5 pairs; GSM8K, CSQA, MMLU, HumanEval & tokens per second, accepted length & $\times$ & $\times$ & $\times$ \\
Speculative cascades \citep{narasimhan2025cascades} & T5, Gemma pairs; translation, summarization, GSM8K, MBPP, QA & batch-one relative latency, rejection rate & $\times$ & $\times$ & $\times$ \\
Ensemble \citep{wang2026diversed} & Llama-3.1, Qwen3, Gemma-3 pairs; GSM8K, CNN/DM, XSum, MBPP & acceptance rate, wall-clock at fixed token count & $\times$ & $\times$ & $\times$ \\
Spec.\ contrastive \citep{yuan2024scd} & Llama-2 7B/70B; WikiText, HumanEval, AlpacaEval, GSM8K & expected acceleration from acceptance rate & $\times$ & $\times$ & $\times$ \\
\citet{xia2026practical} & Qwen3-32B with 0.6B/8B drafters, Qwen3.5 MTP; AIME24, GPQA, LCB & $\laccept$ at several $\Ndraft$, modeled $\crel$, Eq.~\eqref{eq:speedup-relax} & \checkmark & $\times$ & \checkmark \\
\bottomrule
\end{tabular}
\caption{Results reported by the source papers (Appendix~\ref{app:evidence}). Length: completion length reported as a measured quantity; Rounds: verifier rounds or target forward passes reported; Long-form: evaluation on long reasoning traces. (\checkmark) marks indirect or partial coverage: CACTUS reports rejected-token counts, which grow with length and with rounds ending in rejection, on reasoning-mode models, and discusses length only as a caveat and one qualitative case.}
\label{tab:evidence}
\end{table*}

\section{Full Results}
\label{app:results}

Table~\ref{tab:strict} gives the lossless reference of the primary pairs, and Table~\ref{tab:cells} their cells at the most relaxed setting with 95\% paired bootstrap intervals. Results at every tested setting, with plots of completion length, rounds, and accuracy against $\laccept$, are in the released tables.

\begin{table}[t]
  \centering
  \scriptsize
  \setlength{\tabcolsep}{4pt}
  \begin{tabular}{@{}lcccccc@{}}
    \toprule
    Dataset & $n$ & $\laccept^{*}$ & $\Lbar^{*}$ & $R^{*}$ & Acc. & cut off \\
    \midrule
    \multicolumn{7}{@{}l}{\emph{GPT-OSS-20B}} \\
    GSM8K         & 150 & 2.57 & 322     & 94     & 0.960 & 2 \\
    AIME24        & 30  & 2.19 & 8\,717  & 2\,953 & 0.767 & 7 \\
    HumanEval     & 150 & 2.53 & 912     & 272    & 0.960 & 0 \\
    LiveCodeBench & 90  & 2.16 & 3\,456  & 1\,151 & 0.889 & 2 \\
    MT-Bench      & 80  & 2.34 & 1\,264  & 414    & 7.29 & 4 \\
    LongBench-v2  & 150 & 1.91 & 1\,509  & 520    & 0.560 & 1 \\
    \addlinespace[2pt]
    \multicolumn{7}{@{}l}{\emph{Qwen3-8B}} \\
    GSM8K         & 150 & 1.51 & 1\,292  & 510    & 0.800 & 25 \\
    AIME24        & 30  & 1.10 & 19\,141 & 9\,475 & 0.700 & 17 \\
    HumanEval     & 150 & 1.42 & 3\,670  & 1\,535 & 0.827 & 11 \\
    LiveCodeBench & 90  & 1.21 & 8\,029  & 3\,703 & 0.700 & 32 \\
    MT-Bench      & 80  & 1.25 & 2\,073  & 913    & 6.94 & 14 \\
    LongBench-v2  & 150 & 0.70 & 2\,887  & 1\,676 & 0.527 & 3 \\
    \bottomrule
  \end{tabular}
  \caption{Lossless (strict) speculative decoding on the evaluation cases, with no relaxation, for the two primary pairs (Section~\ref{sec:setup}). $\laccept^{*}$: mean accepted draft tokens per verification round; $\Lbar^{*}$: mean completion length in tokens; $R^{*}$: mean verifier rounds; Acc.: pass rate; for MT-Bench, the mean judge score out of 10 (Table~\ref{tab:judge}); cut off: share of runs (\%) that end at the generation budget. Generation budgets (tokens): GSM8K 2\,048, AIME24 32\,768, HumanEval 9\,000, LiveCodeBench 12\,000, MT-Bench 4\,096, LongBench-v2 8\,192. Every comparison in Section~\ref{sec:experiments} is against the lossless run of the same pair, not against autoregressive decoding.}
  \label{tab:strict}
\end{table}

\begin{table}[t]
\centering\scriptsize
\setlength{\tabcolsep}{3.5pt}
\begin{tabular}{@{}l r rr r rr@{}}
\toprule
& & \multicolumn{2}{c}{target prob.\ (median)} & rank & \multicolumn{2}{c}{entropy (mean)} \\
\cmidrule(lr){3-4}\cmidrule(lr){6-7}
Method & only & only & both & only & only & both \\
\midrule
\texttt{mentored\_dec} & 16\% & 0.09 & 0.93 & 37 & 1.8 & 0.8 \\
\texttt{cactus} & 24\% & 0.01 & 0.78 & 234 & 2.1 & 1.1 \\
\texttt{spec\_casc\_opt} & 16\% & 0.07 & 0.92 & 88 & 1.8 & 0.8 \\
\texttt{r\_fuzzy} & 21\% & 0.01 & 0.83 & 134 & 1.8 & 1.0 \\
\texttt{spec\_casc\_tok} & 7\% & 0.29 & 0.99 & 0.9 & 1.5 & 0.5 \\
\bottomrule
\end{tabular}
\caption{What relaxation admits (GPT-OSS-20B, traced runs, loosest settings): committed draft tokens that only the relaxed rule accepts (``only'') against those lossless verification would also accept (``both''). Share of committed draft tokens, median target probability, mean target rank ($0$ is the argmax), mean target entropy (nats).}
\label{tab:admit}
\end{table}

\paragraph{Gentler settings.}
Restricting each method to its gentlest tested setting instead of its loosest gives $\lambda$ with median $1.05$ and mean $1.08$ over the 60 cells of the primary pairs, a median round ratio of $0.98$, and 21 cells that need more rounds than lossless; 27 of the 50 graded cells still lose accuracy (mean drop $0.03$), against 40 ($0.17$) at the loosest. Table~\ref{tab:gentle} (Appendix~\ref{app:pairs}) compares the two settings on all seven pairs run with the matched protocol.

\paragraph{Best accuracy-preserving setting.}
Table~\ref{tab:best} gives, per method and primary pair, how many graded datasets have a tested setting within two accuracy points of lossless and the lowest round ratio among those settings (Section~\ref{sec:analysis}); for \texttt{mentored\_dec} and \texttt{spec\_casc\_tok} every dataset has all four $\alpha$ values, the 41 settings missing from the sweep having been run on the second and third servers against a lossless reference on the same server. For these two rules we reran the setting with the lowest time ratio among those within two points (MT-Bench: with a round ratio below $1$) with a new seed on the second server (some Qwen3-8B arms on the third), paired with lossless at the same seed on the same server: it again has a time ratio below $1$ and accuracy within two points in 15 of the 24 cases (released tables).

\paragraph{MT-Bench.} MT-Bench (Table~\ref{tab:judge}; Appendix~\ref{app:pairs}) separates the rules the same way: across the twelve pairs \texttt{spec\_casc\_tok} scores $-0.09$ to $+1.01$ points against lossless and \texttt{mentored\_dec} $-1.55$ to $+0.21$, while the three aggressive rules lose a median $2.3$ points and up to $4.0$. It is also where relaxation saves the most rounds (median $R/R^{*}$ $0.84$; elsewhere $0.90$ to $0.97$), so the largest saving comes where the aggressive rules lose judged quality.

\begin{table}[t]
\centering\footnotesize
\setlength{\tabcolsep}{3pt}
\begin{tabular}{@{}l cc cc@{}}
\toprule
& \multicolumn{2}{c}{GPT-OSS-20B} & \multicolumn{2}{c}{Qwen3-8B} \\
\cmidrule(lr){2-3}\cmidrule(lr){4-5}
Method & cells & best $R/R^{*}$ & cells & best $R/R^{*}$ \\
\midrule
\texttt{mentored\_dec} & 5/5 & 0.86 & 5/5 & 0.88 \\
\texttt{cactus} & 1/5 & 0.98 & 2/5 & 0.97 \\
\texttt{spec\_casc\_opt} & 2/5 & 0.93 & 2/5 & 0.97 \\
\texttt{r\_fuzzy} & 2/5 & 1.13 & 0/5 & --- \\
\texttt{spec\_casc\_tok} & 5/5 & 0.83 & 5/5 & 0.93 \\
\bottomrule
\end{tabular}
\caption{The best accuracy-preserving setting. For each graded (dataset, method) cell we take, among the method's tested $\alpha$ values, the one with the fewest verifier rounds whose accuracy is within two points of lossless. ``cells'' counts the graded datasets (of five) on which any tested setting qualifies; ``best $R/R^{*}$'' is the lowest round ratio among them. The median best round ratio over the 29 qualifying cells is $0.96$.}
\label{tab:best}
\end{table}

\begin{table}[t]
\centering\scriptsize
\setlength{\tabcolsep}{3pt}
\begin{tabular}{@{}l c l r r@{}}
\toprule
Method & $\alpha$ & score [95\% CI] & no ans. & answered \\
\midrule
\multicolumn{5}{@{}l}{\emph{GPT-OSS-20B}} \\
lossless & --- & 7.29 [6.75,\,7.80] & 0 & 7.29 \\
\texttt{mentored\_dec} & 0.75 & 6.41 [5.73,\,7.09] & 1 & 6.48 \\
\texttt{cactus} & 0.35 & 4.51 [3.70,\,5.34] & 6 & 4.80 \\
\texttt{spec\_casc\_opt} & 0.05 & 5.74 [5.04,\,6.45] & 2 & 5.86 \\
\texttt{r\_fuzzy} & 0.25 & 4.58 [3.94,\,5.23] & 3 & 4.71 \\
\texttt{spec\_casc\_tok} & 0.8 & 7.49 [6.96,\,7.99] & 0 & 7.49 \\
\addlinespace[2pt]
\multicolumn{5}{@{}l}{\emph{Qwen3-8B}} \\
lossless & --- & 6.94 [6.31,\,7.55] & 10 & 7.79 \\
\texttt{mentored\_dec} & 0.75 & 6.40 [5.70,\,7.06] & 13 & 7.45 \\
\texttt{cactus} & 0.35 & 2.99 [2.40,\,3.64] & 22 & 3.74 \\
\texttt{spec\_casc\_opt} & 0.05 & 3.95 [3.43,\,4.49] & 22 & 5.07 \\
\texttt{r\_fuzzy} & 0.25 & 3.05 [2.61,\,3.51] & 20 & 3.73 \\
\texttt{spec\_casc\_tok} & 0.8 & 7.21 [6.59,\,7.80] & 9 & 8.00 \\
\bottomrule
\end{tabular}
\caption{MT-Bench quality of the primary pairs at the loosest settings, judged by Claude (\texttt{claude-fable-5-1}) with FastChat's single-answer grading \citep{zheng2023mtbench} of the first turn (the GPT-4 reference answer is supplied for math, reasoning, and coding questions). score: mean over the 80 questions, 1 to 10, with a bootstrap interval; no ans.: runs that never produced an answer (scored 1 without a judge call); answered: mean over the other runs.}
\label{tab:judge}
\end{table}

\begin{table*}[t]
\centering\scriptsize
\setlength{\tabcolsep}{3pt}
\begin{tabular}{@{}ll rrrr rrrr@{}}
\toprule
& & \multicolumn{4}{c}{GPT-OSS-20B + EAGLE-3 (nebius)} & \multicolumn{4}{c}{Qwen3-8B + EAGLE-3 (RedHatAI)} \\
\cmidrule(lr){3-6}\cmidrule(lr){7-10}
Dataset & Method & $\alpha$ & $\lambda$ [95\% CI] & time [95\% CI] & cut off & $\alpha$ & $\lambda$ [95\% CI] & time [95\% CI] & cut off \\
\midrule
GSM8K & \texttt{mentored\_dec} & 0.75 & 1.16 [1.02,\,1.33] & 0.96 [0.86,\,1.08] & 4 & 0.75 & 1.03 [0.98,\,1.07] & 0.95 [0.91,\,1.00] & 26 \\
 & \texttt{cactus} & 0.35 & 1.61 [1.41,\,1.83] & 1.19 [1.06,\,1.33] & 5 & 0.35 & 1.04 [1.00,\,1.08] & 0.94 [0.90,\,0.97] & 27 \\
 & \texttt{spec\_casc\_opt} & 0.05 & 1.46 [1.28,\,1.65] & 1.18 [1.06,\,1.31] & 5 & 0.05 & 1.33 [1.26,\,1.40] & 1.12 [1.07,\,1.18] & 58 \\
 & \texttt{r\_fuzzy} & 0.25 & 1.70 [1.49,\,1.94] & 1.29 [1.15,\,1.45] & 5 & 0.25 & 0.95 [0.86,\,1.05] & 0.81 [0.74,\,0.90] & 35 \\
 & \texttt{spec\_casc\_tok} & 0.8 & 0.96 [0.88,\,1.04] & 0.91 [0.85,\,0.98] & 1 & 0.8 & 0.99 [0.95,\,1.02] & 0.96 [0.92,\,1.00] & 23 \\
\addlinespace[2pt]
AIME24 & \texttt{mentored\_dec} & 0.75 & 1.69 [1.34,\,2.21] & 1.36 [1.07,\,1.81] & 20 & 0.75 & 1.01 [0.91,\,1.14] & 0.95 [0.83,\,1.10] & 23 \\
 & \texttt{cactus} & 0.18 & 1.78 [1.37,\,2.44] & 1.33 [1.00,\,1.87] & 23 & 0.35 & 1.26 [1.10,\,1.44] & 0.66 [0.56,\,0.77] & 57 \\
 & \texttt{spec\_casc\_opt} & 0.05 & 2.48 [1.89,\,3.45] & 2.01 [1.51,\,2.83] & 43 & 0.05 & 1.42 [1.21,\,1.69] & 1.14 [0.94,\,1.39] & 63 \\
 & \texttt{r\_fuzzy} & 0.25 & 2.13 [1.64,\,2.92] & 1.64 [1.26,\,2.30] & 30 & 0.25 & 1.29 [1.08,\,1.53] & 1.18 [0.96,\,1.45] & 37 \\
 & \texttt{spec\_casc\_tok} & 0.8 & 1.48 [1.09,\,2.04] & 1.33 [0.96,\,1.85] & 20 & 0.8 & 0.98 [0.86,\,1.11] & 0.93 [0.81,\,1.07] & 13 \\
\addlinespace[2pt]
HumanEval & \texttt{mentored\_dec} & 0.75 & 1.30 [1.15,\,1.46] & 1.09 [0.97,\,1.21] & 1 & 0.75 & 1.08 [1.01,\,1.16] & 1.02 [0.95,\,1.09] & 13 \\
 & \texttt{cactus} & 0.35 & 1.79 [1.61,\,1.99] & 1.35 [1.23,\,1.49] & 1 & 0.35 & 1.14 [1.07,\,1.23] & 1.01 [0.94,\,1.09] & 20 \\
 & \texttt{spec\_casc\_opt} & 0.05 & 1.73 [1.52,\,1.97] & 1.43 [1.28,\,1.60] & 2 & 0.05 & 1.72 [1.57,\,1.89] & 1.46 [1.34,\,1.60] & 40 \\
 & \texttt{r\_fuzzy} & 0.25 & 1.64 [1.46,\,1.84] & 1.32 [1.18,\,1.47] & 1 & 0.25 & 1.65 [1.50,\,1.82] & 1.41 [1.28,\,1.56] & 38 \\
 & \texttt{spec\_casc\_tok} & 0.8 & 1.05 [0.96,\,1.15] & 0.98 [0.91,\,1.07] & 0 & 0.8 & 1.08 [1.02,\,1.14] & 1.05 [0.99,\,1.11] & 11 \\
\addlinespace[2pt]
LiveCodeBench & \texttt{mentored\_dec} & 0.75 & 1.25 [1.14,\,1.36] & 1.01 [0.93,\,1.11] & 9 & 0.75 & 1.05 [1.01,\,1.09] & 0.95 [0.92,\,0.99] & 34 \\
 & \texttt{cactus} & 0.18 & 1.54 [1.33,\,1.77] & 1.14 [0.99,\,1.32] & 11 & 0.35 & 1.15 [1.08,\,1.22] & 0.82 [0.76,\,0.89] & 54 \\
 & \texttt{spec\_casc\_opt} & 0.05 & 1.64 [1.46,\,1.84] & 1.32 [1.17,\,1.48] & 19 & 0.05 & 1.27 [1.20,\,1.36] & 1.09 [1.02,\,1.16] & 61 \\
 & \texttt{r\_fuzzy} & 0.25 & 1.65 [1.45,\,1.89] & 1.27 [1.12,\,1.45] & 16 & 0.25 & 1.07 [0.96,\,1.20] & 0.92 [0.82,\,1.03] & 61 \\
 & \texttt{spec\_casc\_tok} & 0.8 & 1.16 [1.03,\,1.31] & 1.04 [0.92,\,1.17] & 7 & 0.8 & 1.00 [0.97,\,1.04] & 0.96 [0.92,\,0.99] & 30 \\
\addlinespace[2pt]
MT-Bench & \texttt{mentored\_dec} & 0.75 & 1.10 [0.98,\,1.22] & 0.87 [0.78,\,0.97] & 1 & 0.75 & 1.04 [0.98,\,1.11] & 0.93 [0.87,\,0.99] & 21 \\
 & \texttt{cactus} & 0.35 & 1.18 [1.04,\,1.35] & 0.82 [0.72,\,0.95] & 6 & 0.35 & 1.16 [1.03,\,1.29] & 0.75 [0.68,\,0.84] & 32 \\
 & \texttt{spec\_casc\_opt} & 0.05 & 1.09 [0.98,\,1.21] & 0.87 [0.78,\,0.97] & 2 & 0.05 & 1.22 [1.14,\,1.32] & 1.01 [0.95,\,1.09] & 28 \\
 & \texttt{r\_fuzzy} & 0.25 & 1.20 [1.07,\,1.35] & 0.92 [0.81,\,1.04] & 4 & 0.25 & 1.12 [1.04,\,1.21] & 0.92 [0.85,\,0.99] & 26 \\
 & \texttt{spec\_casc\_tok} & 0.8 & 0.96 [0.88,\,1.05] & 0.90 [0.84,\,0.98] & 1 & 0.8 & 1.04 [0.97,\,1.11] & 0.99 [0.93,\,1.05] & 14 \\
\addlinespace[2pt]
LongBench-v2 & \texttt{mentored\_dec} & 0.75 & 1.59 [1.38,\,1.83] & 1.04 [1.02,\,1.07] & 3 & 0.75 & 1.06 [0.99,\,1.14] & 0.99 [0.92,\,1.07] & 5 \\
 & \texttt{cactus} & 0.35 & 1.77 [1.52,\,2.05] & 1.03 [1.00,\,1.05] & 2 & 0.35 & 2.10 [1.86,\,2.37] & 1.05 [0.94,\,1.16] & 57 \\
 & \texttt{spec\_casc\_opt} & 0.05 & 1.79 [1.54,\,2.07] & 1.07 [1.04,\,1.10] & 14 & 0.05 & 1.16 [1.05,\,1.28] & 0.90 [0.83,\,0.98] & 17 \\
 & \texttt{r\_fuzzy} & 0.25 & 1.69 [1.45,\,1.98] & 1.05 [1.02,\,1.07] & 5 & 0.25 & 1.08 [0.99,\,1.18] & 0.99 [0.90,\,1.07] & 11 \\
 & \texttt{spec\_casc\_tok} & 0.8 & 1.13 [0.98,\,1.29] & 1.00 [0.98,\,1.02] & 1 & 0.8 & 1.00 [0.92,\,1.07] & 0.97 [0.90,\,1.04] & 4 \\
\bottomrule
\end{tabular}
\caption{The primary pairs' cells at each method's most relaxed tested setting: $\alpha$; length ratio $\lambda$ and per-request time ratio to lossless, each with its 95\% paired bootstrap interval over cases (10\,000 resamples); cut off: share of runs (\%) that end at the generation budget (lossless: Table~\ref{tab:strict}). Round ratios: Table~\ref{tab:rounds}; accuracy: Table~\ref{tab:length}; acceptance gains and round-ratio intervals: released tables.}
\label{tab:cells}
\end{table*}

\section{Seeds, Draft Length, Temperature, and Hardware}
\label{app:seeds}

\paragraph{Hardware and compute.}
Seed 0 ran on an H100 PCIe server (the first server), seeds 1 and 2 on an H100 SXM server (the second); the Qwen3-8B additions of Appendices~\ref{app:results} and~\ref{app:speedbench} (except the first 40 SPEED-Bench prompts per arm, run on the second), the Qwen3-8B controls and AIME24 seeds of this appendix, and the pairs of Appendix~\ref{app:pairs} ran on a third server, also H100 SXM. On the first server, the first cases of each benchmark ran right after a fresh server start and the rest on a process reused across requests. On the 28 GPT-OSS-20B cells with three seeds, $\lambda$ and the round ratio replicate across the first two servers (mean $\lambda$ $1.49$, $1.44$, $1.48$ and mean round ratio $1.04$, $1.01$, $1.03$ for seeds 0, 1, 2); the time ratio does not ($1.17$, $1.01$, $1.03$): it is lower on the second server in 23 and 24 of the 28 cells. Fitting per-run time per round as $c_0 + c_1 \times$ (tokens committed per round) locates the difference. On the first server $c_0$ is $12$ to $14$\,ms and $c_1$ is $2.5$ to $3.2$\,ms per token on HumanEval, MT-Bench, LiveCodeBench, and AIME24 ($R^2$ $0.39$ to $0.87$; GSM8K rounds are too few and too short to fit, and LongBench-v2 is dominated by the prefill), so a relaxed arm's time per round is $1.12\times$ lossless on average and above $1.05\times$ in 26 of the 30 GPT-OSS-20B cells. On the second server $c_0$ is $7.1$ to $7.6$\,ms and $c_1$ at most $0.08$\,ms per token ($R^2 \le 0.13$), and the time-per-round ratio is $1.00$: at seeds 1 and 2, 14 and 12 cells save rounds, the same cells save time, and the round and time ratios are never more than $0.06$ apart. The third server's nodes are not uniform (one node ran rounds about $11\%$ slower than its neighbours on one day and not on the next), so every Qwen3-8B time ratio from that server is reported over the case pairs whose arm and lossless reference ran on the same node, or marked when fewer than ten pairs did; round ratios are unaffected. The experiments used about 750 H100 GPU-hours: 441 logged on the second and third servers and about 300 on the first.

\paragraph{Seeds.}
To test whether the paired bootstrap over cases captures run-to-run variation, we repeated the lossless reference on GPT-OSS-20B with seeds 1 and 2 on five datasets: on every dataset the three runs' mean completion lengths lie inside the 95\% bootstrap interval of the seed-0 mean, and $\laccept^{*}$ moves by at most $0.10$. We then repeated 28 of the 30 GPT-OSS-20B loosest-setting cells and 22 of the 30 Qwen3-8B cells, with their lossless reference, with seeds 1 and 2 on the second server (Qwen3-8B AIME24 on the third; per-seed values in the released tables). On GPT-OSS-20B, $\lambda$ exceeds $1$ in all three seeds in 24 cells; the four exceptions are all \texttt{spec\_casc\_tok}, whose $\lambda$ crosses $1$ between seeds. The seed-to-seed standard deviation has median $0.08$ for $\lambda$ (maximum $0.27$, on AIME24 with 30 cases), $0.06$ for the round ratio, and $0.09$ for the time ratio. The seed-0 interval of $\lambda$ contains both new seeds in 21 of the 28 cells; five of the seven exceptions miss it by at most $0.04$ and the other two are on AIME24. In 20 cells the round ratio lies on the same side of $1$ in all three seeds; the eight that cross it are three AIME24 cells and five cells within $0.13$ of $1$ in every seed. On Qwen3-8B the seeds agree more closely: the standard deviation has median $0.03$ for $\lambda$ (maximum $0.10$), $0.03$ for the round ratio, and $0.04$ for the time ratio; $\lambda$ exceeds $1$ in all three seeds in 16 of the 22 cells, the six exceptions being cells whose seed-0 $\lambda$ is at most $1.12$; and the round ratio lies on the same side of $1$ in all three seeds in 21 cells (the exception, HumanEval \texttt{spec\_casc\_tok}, stays within $0.04$ of $1$). The GPT-OSS-20B LongBench-v2 cells of \texttt{cactus} and \texttt{spec\_casc\_tok} and the Qwen3-8B LongBench-v2 cells were run once; the Qwen3-8B AIME24 cells of \texttt{mentored\_dec}, \texttt{cactus}, and \texttt{spec\_casc\_tok} have five seeds instead (below).

\paragraph{AIME24 over five seeds.}
Because AIME24 has 30 cases, we repeated its GPT-OSS-20B arms with seeds 1 to 4 on the second server. Accuracy over the five seeds, with a two-level bootstrap interval over problems and seeds: lossless $0.79$ $[0.65, 0.91]$; \texttt{mentored\_dec} $0.68$ $[0.53, 0.82]$; \texttt{spec\_casc\_tok} $0.69$ $[0.54, 0.82]$; \texttt{cactus} $0.51$ $[0.36, 0.67]$; \texttt{r\_fuzzy} $0.48$ $[0.33, 0.63]$; \texttt{spec\_casc\_opt} $0.39$ $[0.25, 0.53]$. The seed-to-seed standard deviation is $0.04$ to $0.06$, one to two problems. Every rule scores below lossless on every seed: the three aggressive rules by 27 to 39 points, \texttt{mentored\_dec} and \texttt{spec\_casc\_tok} by 11 and 10 points with overlapping intervals. On 30 problems $\lambda$ varies widely across seeds (\texttt{spec\_casc\_tok} $0.96$ to $1.60$, \texttt{mentored\_dec} $1.28$ to $2.15$), so the single-seed AIME24 values of Table~\ref{tab:length} carry that spread. On Qwen3-8B (seeds 1 to 4 on the third server) the same five seeds give lossless $0.72$ $[0.57, 0.85]$; \texttt{mentored\_dec} $0.72$ $[0.56, 0.87]$; \texttt{spec\_casc\_tok} $0.73$ $[0.58, 0.86]$; \texttt{cactus} $0.21$ $[0.09, 0.35]$; \texttt{spec\_casc\_opt} $0.32$ $[0.19, 0.45]$; \texttt{r\_fuzzy} $0.35$ $[0.21, 0.50]$: the two gentle rules match lossless on average and the three aggressive rules lose 37 to 51 points.

\paragraph{Draft length.}
We ran lossless decoding at $\Ndraft \in \{2, 3, 4, 6, 8, 10\}$ on GSM8K and LiveCodeBench, paired with $\Ndraft = 6$ on the same server and cases (GPT-OSS-20B on the second server, Qwen3-8B on the third; released tables). On GPT-OSS-20B shorter drafts save $5$ to $9\%$ of the time (every interval touches $1$), longer drafts cost $6$ to $24\%$, completion length is unchanged ($\lambda$ $0.96$ to $1.06$), and accuracy moves by at most five points, within the binomial error of the case counts. On Qwen3-8B, whose head accepts fewer tokens per round ($\laccept^{*}$ $1.50$ and $1.13$ at $\Ndraft = 6$), shorter drafts change the time by $-10\%$ to $+1\%$ and longer ones cost $8$ to $16\%$, with $\lambda$ $0.99$ to $1.03$. The reference of the body is therefore within $9\%$ and $10\%$ of the fastest lossless draft length on the two primary pairs, and no lossless draft length changes the length distribution.

\paragraph{Temperature.}
We ran lossless decoding on GPT-OSS-20B at sampling temperature $1.2$ and $1.5$ on the second server, paired with $T = 1.0$ on the same server and cases (released tables). At $T = 1.2$, GSM8K completions are $1.27\times$ longer ($[1.10, 1.47]$), rounds $1.37\times$ and time $1.35\times$, with accuracy unchanged ($0.96$ against $0.97$); on LiveCodeBench $\lambda = 1.07$ ($[0.99, 1.17]$) and time $1.16\times$. At $T = 1.5$, GSM8K inflates by $3.94\times$ with $48\%$ of runs at the budget and accuracy $0.31$, and LiveCodeBench by $1.82\times$ with accuracy $0.08$. Acceptance falls with temperature ($\laccept^{*}$ $2.58$, $2.36$, $1.79$ on GSM8K), so a hotter lossless sampler pays in length and in rounds per token at once. A change of $+0.2$ in temperature inflates GSM8K about as much as \texttt{mentored\_dec} at its loosest setting does at $T = 1.0$ ($\lambda = 1.16$), without any relaxation. Qwen3-8B behaves differently: its completions barely lengthen at $T = 1.2$ or $1.5$ ($\lambda$ $1.02$ to $1.04$, accuracy within three points on GSM8K and down nine on LiveCodeBench at $T = 1.5$), but its acceptance falls faster ($\laccept^{*}$ $1.50$, $1.28$, $0.82$ on GSM8K), so rounds and time still rise by $12$ to $49\%$.

\paragraph{Qwen3-8B at its recommended sampler.}
The body runs every target at temperature $1.0$ and top-$p$ $1.0$. Qwen3-8B's bundled sampling defaults are overridden so that every target uses the same sampler, and its context window is extended to 65\,536 tokens with YaRN \citep{peng2024yarn} (factor $1.6$) to match GPT-OSS-20B. Table~\ref{tab:t06} repeats the five rules at their loosest settings on Qwen3-8B with its recommended thinking-mode sampler (temperature $0.6$, top-$p$ $0.95$, top-$k$ 20) on GSM8K and LiveCodeBench, on the third server, paired with lossless at the same sampler. Lossless acceptance rises ($\laccept^{*}$ $1.68$ and $1.36$ against $1.50$ and $1.13$ at temperature $1.0$ on the same server) and the picture changes for the gentle rules: \texttt{mentored\_dec}, \texttt{cactus}, and \texttt{spec\_casc\_tok} no longer lengthen completions ($\lambda$ $1.01$ to $1.04$), save $4$ to $7\%$ of the rounds, and keep accuracy. \texttt{r\_fuzzy} keeps the length but loses $15$ and $57$ accuracy points, and \texttt{spec\_casc\_opt} inflates by $1.52\times$ and $1.48\times$ with $87\%$ and $99\%$ of its runs at the budget and accuracy $0.20$ and $0.00$; its round ratio of $0.75$ on LiveCodeBench is an acceptance gain of $2.0\times$ against an inflation that the budget cuts short, not a usable setting.

\begin{table*}[t]
\centering\scriptsize
\setlength{\tabcolsep}{5pt}
\begin{tabular}{@{}ll r l l r r@{}}
\toprule
Dataset & Method & $\laccept$ & $\lambda$ [95\% CI] & $R/R^{*}$ [95\% CI] & Acc. & cut off \\
\midrule
GSM8K & \texttt{mentored\_dec} & 1.85 & 1.01 [0.97,\,1.04] & 0.94 [0.91,\,0.98] & 0.81 & 23\% \\
 & \texttt{cactus} & 1.81 & 1.01 [0.97,\,1.05] & 0.96 [0.92,\,1.00] & 0.81 & 23\% \\
 & \texttt{spec\_casc\_opt} & 3.34 & 1.52 [1.43,\,1.62] & 0.95 [0.89,\,1.02] & 0.20 & 87\% \\
 & \texttt{r\_fuzzy} & 2.05 & 1.01 [0.92,\,1.09] & 0.89 [0.81,\,0.96] & 0.64 & 39\% \\
 & \texttt{spec\_casc\_tok} & 1.80 & 1.01 [0.97,\,1.04] & 0.96 [0.92,\,1.00] & 0.78 & 26\% \\
\addlinespace[2pt]
LiveCodeBench & \texttt{mentored\_dec} & 1.57 & 1.04 [1.00,\,1.09] & 0.94 [0.90,\,0.99] & 0.68 & 34\% \\
 & \texttt{cactus} & 1.53 & 1.01 [0.98,\,1.04] & 0.93 [0.90,\,0.97] & 0.72 & 29\% \\
 & \texttt{spec\_casc\_opt} & 3.73 & 1.48 [1.36,\,1.63] & 0.75 [0.68,\,0.83] & 0.00 & 99\% \\
 & \texttt{r\_fuzzy} & 1.70 & 1.08 [0.98,\,1.19] & 0.97 [0.87,\,1.07] & 0.10 & 52\% \\
 & \texttt{spec\_casc\_tok} & 1.51 & 1.02 [0.98,\,1.06] & 0.95 [0.91,\,0.99] & 0.68 & 32\% \\
\bottomrule
\end{tabular}
\caption{Qwen3-8B at temperature $0.6$, top-$p$ $0.95$, top-$k$ 20 (seed 0, third server), loosest settings, against lossless at the same sampler and cases: lossless $\laccept^{*}$ $1.68$ (GSM8K) and $1.36$ (LiveCodeBench), accuracy $0.79$ and $0.67$, cut off $25\%$ and $33\%$.}
\label{tab:t06}
\end{table*}

\section{Other Targets and Drafters}
\label{app:pairs}

Table~\ref{tab:pairs} summarizes the twelve pairs. Tables~\ref{tab:length-all} and~\ref{tab:rounds-all} list their 360 cells at each rule's most relaxed tested setting, Table~\ref{tab:strict-all} gives the lossless references of the ten pairs beyond the primary ones, and Table~\ref{tab:gentle} compares the gentlest and the most relaxed tested setting on the seven pairs run with the matched protocol. The released tables give every tested setting of every cell with $\laccept$, the cut-off share, the wall-clock ratio, and 95\% intervals.

The targets are \nolinkurl{openai/gpt-oss-20b}, \nolinkurl{Qwen/Qwen3-8B}, \nolinkurl{meta-llama/Llama-3.1-8B-Instruct}, and \nolinkurl{deepseek-ai/DeepSeek-R1-Distill-Llama-8B}. The drafters beyond the primary ones are \nolinkurl{RedHatAI/gpt-oss-20b-speculator.eagle3} for GPT-OSS-20B; \nolinkurl{RedHatAI/Qwen3-8B-speculator.peagle}, \nolinkurl{deepseek-ai/dspark_qwen3_8b_block7}, \nolinkurl{Qwen/Qwen3-0.6B}, and \nolinkurl{Qwen/Qwen3-1.7B} for Qwen3-8B; \nolinkurl{yuhuili/EAGLE3-LLaMA3.1-Instruct-8B}, \nolinkurl{yuhuili/EAGLE-LLaMA3.1-Instruct-8B}, and \nolinkurl{meta-llama/Llama-3.2-1B-Instruct} for Llama-3.1-8B-Instruct; and \nolinkurl{yuhuili/EAGLE3-DeepSeek-R1-Distill-LLaMA-8B} and \nolinkurl{meta-llama/Llama-3.2-1B-Instruct} for DeepSeek-R1-Distill-Llama-8B. The standalone drafters run through vLLM's draft-model path. The dedicated drafters other than P-EAGLE ran the matched protocol of Section~\ref{sec:protocol}; P-EAGLE and the standalone drafters ran the loosest setting of each rule. These ten pairs ran on the third server, one persistent vLLM process per arm, each arm paired case by case with its own pair's lossless run. The Llama-3.1-8B-Instruct and Llama-3.2-1B-Instruct weights come from public mirrors byte-identical to Meta's files, and the Llama models' bundled sampler defaults are overridden as for Qwen3-8B.

The published EAGLE heads for the Llama models cap their position embeddings at 2\,048; we extend the configuration to 65\,536 positions, and do the same on LongBench-v2 for DSpark, P-EAGLE, and the standalone Qwen3 drafters, whose configurations stop at 40\,960 (identical below it). The head for R1-Distill-Llama-8B stops drafting usefully beyond about 2\,048 positions of a trace (lossless $\laccept$ $2.2$ to $3.4$ within the first 2\,048 positions, $0.2$ to $1.2$ beyond), which is why its lossless $\laccept^{*}$ on LiveCodeBench, AIME24, and LongBench-v2 is below $1$; the Llama-3.2-1B drafter shares R1-Distill's vocabulary except for seven special tokens. On LongBench-v2, whose prompts run 10\,000 to 45\,000 tokens, the second GPT-OSS-20B head and Llama-3.1-8B-Instruct's EAGLE-3 head also nearly stop drafting (lossless $\laccept^{*}$ $0.05$ and $0.06$; R1-Distill's $0.03$); there the rules either change little or, \texttt{cactus} above all, produce runaway answers, and we report these cells as measured. Relaxation raises $\laccept$ in 347 of the 360 cells and lowers it in 13, 12 of them \texttt{r\_fuzzy} and 8 on the Llama EAGLE heads (gain $0.80$ to $0.995$). HumanEval is graded by executing the last code block of the answer, as for the primary pairs; Llama-3.1-8B-Instruct often ends its answer with a usage example, so its HumanEval accuracies are lower bounds (executing instead the last block that defines the function raises lossless accuracy by at most $4$ points and a relaxed arm's by at most $10$, and changes no conclusion). A Medusa head for Llama-3.1-8B-Instruct serves in vLLM but exposes no draft distribution to the verifier, so four of the five rules reduce to lossless verification on it, and it is not reported.

\begin{table*}[t]
\centering\scriptsize
\renewcommand{\arraystretch}{0.9}
\setlength{\tabcolsep}{2.6pt}
\begin{tabular}{@{}l ccccc ccccc@{}}
\toprule
Dataset & mentored & cactus & casc-opt & r-fuzzy & casc-tok & mentored & cactus & casc-opt & r-fuzzy & casc-tok \\
\midrule
 & \multicolumn{5}{c}{GPT-OSS-20B + EAGLE-3 (nebius)} & \multicolumn{5}{c}{Qwen3-8B + EAGLE-3 (RedHatAI)} \\
\cmidrule(lr){2-6}\cmidrule(lr){7-11}
GSM8K & 1.16 (.93) & 1.61 (.91) & 1.46 (.94) & 1.70 (.87) & 0.96$^{\circ}$ (.97) & 1.03$^{\circ}$ (.77) & 1.04$^{\circ}$ (.77) & 1.33 (.49) & 0.95$^{\circ}$ (.55) & 0.99$^{\circ}$ (.79) \\
AIME24 & 1.69 (.63) & 1.78 (.60) & 2.48 (.37) & 2.13 (.50) & 1.48 (.70) & 1.01$^{\circ}$ (.73) & 1.26 (.23) & 1.42 (.30) & 1.29 (.40) & 0.98$^{\circ}$ (.70) \\
HumanEval & 1.30 (.96) & 1.79 (.87) & 1.73 (.82) & 1.64 (.63) & 1.05$^{\circ}$ (.94) & 1.08 (.85) & 1.14 (.75) & 1.72 (.49) & 1.65 (.16) & 1.08 (.85) \\
LiveCodeBench & 1.25 (.84) & 1.54 (.61) & 1.64 (.49) & 1.65 (.34) & 1.16 (.88) & 1.05 (.63) & 1.15 (.44) & 1.27 (.33) & 1.07$^{\circ}$ (.02) & 1.00$^{\circ}$ (.71) \\
MT-Bench & 1.10$^{\circ}$ (6.41) & 1.18 (4.51) & 1.09$^{\circ}$ (5.74) & 1.20 (4.58) & 0.96$^{\circ}$ (7.49) & 1.04$^{\circ}$ (6.40) & 1.16 (2.99) & 1.22 (3.95) & 1.12 (3.05) & 1.04$^{\circ}$ (7.21) \\
LongBench-v2 & 1.59 (.51) & 1.77 (.36) & 1.79 (.47) & 1.69 (.40) & 1.13$^{\circ}$ (.53) & 1.06$^{\circ}$ (.51) & 2.10 (.29) & 1.16 (.37) & 1.08$^{\circ}$ (.44) & 1.00$^{\circ}$ (.49) \\
\addlinespace[1.5pt]
 & \multicolumn{5}{c}{GPT-OSS-20B + EAGLE-3 (RedHatAI)} & \multicolumn{5}{c}{Qwen3-8B + P-EAGLE (RedHatAI)} \\
\cmidrule(lr){2-6}\cmidrule(lr){7-11}
GSM8K & 1.31 (.95) & 2.49 (.83) & 1.35 (.93) & 2.40 (.78) & 0.87 (.97) & 1.02$^{\circ}$ (.79) & 1.04 (.77) & 1.02$^{\circ}$ (.77) & 1.28 (.51) & 1.00$^{\circ}$ (.73) \\
AIME24 & 1.24$^{\circ}$ (.80) & 1.35$^{\circ}$ (.20) & 1.44 (.63) & 1.68 (.60) & 1.28 (.77) & 1.04$^{\circ}$ (.70) & 1.20 (.60) & 1.08$^{\circ}$ (.67) & 1.50 (.37) & 0.98$^{\circ}$ (.73) \\
HumanEval & 1.48 (.95) & 3.03 (.63) & 1.62 (.84) & 3.11 (.23) & 1.07$^{\circ}$ (.97) & 1.03$^{\circ}$ (.80) & 1.06 (.78) & 1.14 (.71) & 1.67 (.25) & 1.00$^{\circ}$ (.81) \\
LiveCodeBench & 1.43 (.81) & 2.55 (.16) & 1.46 (.73) & 1.28 (.39) & 1.09$^{\circ}$ (.91) & 1.01$^{\circ}$ (.67) & 1.10 (.56) & 1.04$^{\circ}$ (.51) & 1.31 (.03) & 1.04$^{\circ}$ (.64) \\
MT-Bench & 1.36 (6.36) & 1.56 (3.52) & 1.31 (5.40) & 1.58 (3.29) & 1.11$^{\circ}$ (7.56) & 1.05$^{\circ}$ (6.33) & 1.14 (4.14) & 1.02$^{\circ}$ (5.38) & 1.22 (3.35) & 1.02$^{\circ}$ (6.88) \\
LongBench-v2 & 1.07$^{\circ}$ (.57) & 2.92 (.24) & 1.01$^{\circ}$ (.49) & 0.98$^{\circ}$ (.57) & 1.02$^{\circ}$ (.51) & 1.04$^{\circ}$ (.51) & 1.07$^{\circ}$ (.47) & 1.10 (.48) & 1.14 (.45) & 0.99$^{\circ}$ (.47) \\
\addlinespace[1.5pt]
 & \multicolumn{5}{c}{Qwen3-8B + DSpark (DeepSeek)} & \multicolumn{5}{c}{Qwen3-8B + Qwen3-0.6B} \\
\cmidrule(lr){2-6}\cmidrule(lr){7-11}
GSM8K & 0.94 (.81) & 0.93 (.79) & 1.10 (.64) & 1.04$^{\circ}$ (.69) & 0.91 (.83) & 0.94 (.83) & 0.93 (.83) & 0.98$^{\circ}$ (.79) & 1.01$^{\circ}$ (.76) & 0.95 (.83) \\
AIME24 & 1.13 (.63) & 1.09$^{\circ}$ (.70) & 1.52 (.37) & 1.20 (.60) & 1.10$^{\circ}$ (.67) & 0.98$^{\circ}$ (.70) & 0.91$^{\circ}$ (.77) & 0.95$^{\circ}$ (.70) & 1.06$^{\circ}$ (.50) & 0.97$^{\circ}$ (.67) \\
HumanEval & 1.10 (.80) & 1.11 (.82) & 1.43 (.61) & 1.24 (.64) & 1.05$^{\circ}$ (.81) & 0.92 (.87) & 0.94$^{\circ}$ (.87) & 0.95$^{\circ}$ (.83) & 1.03$^{\circ}$ (.85) & 0.95$^{\circ}$ (.87) \\
LiveCodeBench & 1.02$^{\circ}$ (.66) & 1.02$^{\circ}$ (.68) & 1.04 (.62) & 1.07 (.41) & 1.01$^{\circ}$ (.68) & 0.92 (.72) & 0.95 (.76) & 0.95 (.70) & 1.02$^{\circ}$ (.61) & 0.94 (.79) \\
MT-Bench & 1.01$^{\circ}$ (6.67) & 1.06$^{\circ}$ (5.81) & 1.16 (4.47) & 1.06$^{\circ}$ (5.35) & 0.96$^{\circ}$ (6.92) & 0.94 (6.95) & 0.91 (6.33) & 0.87 (6.74) & 0.89 (6.10) & 0.97$^{\circ}$ (7.17) \\
LongBench-v2 & 0.95$^{\circ}$ (.50) & 0.97$^{\circ}$ (.47) & 1.10 (.42) & 1.06$^{\circ}$ (.39) & 0.94$^{\circ}$ (.50) & 0.83 (.44) & 0.76 (.43) & 0.80 (.45) & 0.81 (.39) & 0.85 (.51) \\
\addlinespace[1.5pt]
 & \multicolumn{5}{c}{Qwen3-8B + Qwen3-1.7B} & \multicolumn{5}{c}{Llama-3.1-8B-Instruct + EAGLE-3 (yuhuili)} \\
\cmidrule(lr){2-6}\cmidrule(lr){7-11}
GSM8K & 0.93 (.83) & 0.93 (.83) & 0.96 (.82) & 0.92 (.85) & 0.94 (.81) & 1.59 (.69) & 5.48 (.16) & 1.58 (.48) & 1.21$^{\circ}$ (.69) & 0.83$^{\circ}$ (.83) \\
AIME24 & 0.83 (.80) & 0.85 (.70) & 0.92$^{\circ}$ (.73) & 0.79 (.77) & 0.89$^{\circ}$ (.80) & 2.86 (.03) & 4.10 (.00) & 0.81$^{\circ}$ (.00) & 1.06$^{\circ}$ (.03) & 0.40 (.00) \\
HumanEval & 0.91 (.84) & 0.89 (.83) & 0.92 (.85) & 0.94$^{\circ}$ (.82) & 0.92 (.85) & 1.28 (.45) & 13.23 (.05) & 1.90 (.35) & 2.77 (.27) & 1.02$^{\circ}$ (.53) \\
LiveCodeBench & 0.91 (.73) & 0.91 (.77) & 0.92 (.80) & 0.92 (.76) & 0.93 (.71) & 1.03$^{\circ}$ (.14) & 12.77 (.00) & 2.58 (.02) & 1.68 (.10) & 0.87$^{\circ}$ (.17) \\
MT-Bench & 0.91 (7.01) & 0.92 (6.80) & 0.93 (6.66) & 0.90 (7.33) & 0.93 (7.09) & 1.71 (4.53) & 4.29 (2.00) & 1.08$^{\circ}$ (3.09) & 3.73 (2.77) & 0.84$^{\circ}$ (6.42) \\
LongBench-v2 & 0.83 (.51) & 0.81 (.44) & 0.80 (.33) & 0.85 (.44) & 0.86 (.42) & 3.18 (.37) & 13.22 (.25) & 8.26 (.27) & 2.44 (.32) & 1.06$^{\circ}$ (.37) \\
\addlinespace[1.5pt]
 & \multicolumn{5}{c}{Llama-3.1-8B-Instruct + EAGLE-1 (yuhuili)} & \multicolumn{5}{c}{Llama-3.1-8B-Instruct + Llama-3.2-1B} \\
\cmidrule(lr){2-6}\cmidrule(lr){7-11}
GSM8K & 1.63 (.72) & 5.84 (.06) & 2.87 (.35) & 3.59 (.26) & 0.86$^{\circ}$ (.79) & 0.86$^{\circ}$ (.75) & 1.44 (.60) & 0.81 (.72) & 1.17$^{\circ}$ (.61) & 0.83 (.86) \\
AIME24 & 2.08 (.03) & 4.41 (.00) & 2.70 (.00) & 1.10$^{\circ}$ (.00) & 0.46 (.03) & 0.35 (.03) & 0.56 (.00) & 0.30 (.07) & 0.55 (.00) & 1.07$^{\circ}$ (.03) \\
HumanEval & 1.03$^{\circ}$ (.53) & 9.37 (.01) & 4.88 (.08) & 1.39 (.49) & 1.03$^{\circ}$ (.54) & 1.05$^{\circ}$ (.40) & 1.15 (.38) & 1.07 (.41) & 1.20 (.23) & 0.94 (.57) \\
LiveCodeBench & 3.43 (.10) & 4.05 (.00) & 3.52 (.02) & 3.86 (.01) & 0.80 (.11) & 1.01$^{\circ}$ (.07) & 1.49 (.01) & 0.91$^{\circ}$ (.13) & 1.37 (.01) & 0.79 (.22) \\
MT-Bench & 2.13 (4.40) & 5.60 (1.96) & 1.88 (2.29) & 4.35 (2.11) & 0.83 (6.21) & 1.35$^{\circ}$ (5.34) & 2.45 (3.86) & 1.07$^{\circ}$ (5.14) & 1.60 (4.13) & 0.92$^{\circ}$ (6.54) \\
LongBench-v2 & 1.41$^{\circ}$ (.35) & 8.13 (.22) & 4.24 (.21) & 1.19$^{\circ}$ (.36) & 0.73$^{\circ}$ (.36) & 1.06$^{\circ}$ (.33) & 2.45 (.24) & 0.85$^{\circ}$ (.29) & 1.93 (.21) & 0.59 (.36) \\
\addlinespace[1.5pt]
 & \multicolumn{5}{c}{R1-Distill-Llama-8B + EAGLE-3 (yuhuili)} & \multicolumn{5}{c}{R1-Distill-Llama-8B + Llama-3.2-1B} \\
\cmidrule(lr){2-6}\cmidrule(lr){7-11}
GSM8K & 1.11 (.69) & 1.46 (.55) & 1.11 (.68) & 1.51 (.47) & 1.01$^{\circ}$ (.75) & 1.02$^{\circ}$ (.73) & 2.21 (.44) & 3.17 (.51) & 2.03 (.49) & 0.97$^{\circ}$ (.77) \\
AIME24 & 0.87 (.33) & 2.60 (.00) & 1.32 (.13) & 0.84 (.37) & 0.96$^{\circ}$ (.27) & 1.00$^{\circ}$ (.20) & 2.43 (.00) & 1.45 (.17) & 0.66 (.03) & 1.37 (.37) \\
HumanEval & 1.08 (.84) & 1.97 (.16) & 1.28 (.59) & 1.19 (.51) & 1.02$^{\circ}$ (.81) & 1.26 (.76) & 1.88 (.26) & 1.40 (.64) & 1.51 (.45) & 1.00$^{\circ}$ (.77) \\
LiveCodeBench & 1.07 (.43) & 1.33 (.02) & 1.03$^{\circ}$ (.38) & 1.12 (.17) & 0.95 (.59) & 1.18 (.32) & 1.28 (.04) & 1.09 (.30) & 0.88 (.14) & 1.04$^{\circ}$ (.42) \\
MT-Bench & 1.09 (5.09) & 1.31 (2.82) & 1.14 (4.83) & 1.51 (2.99) & 0.94$^{\circ}$ (5.39) & 1.02$^{\circ}$ (4.95) & 2.42 (1.90) & 1.58 (3.50) & 1.52 (3.03) & 0.92$^{\circ}$ (5.62) \\
LongBench-v2 & 1.03$^{\circ}$ (.35) & 4.00 (.29) & 1.23 (.33) & 0.99$^{\circ}$ (.29) & 1.01$^{\circ}$ (.32) & 0.95$^{\circ}$ (.31) & 2.64 (.25) & 3.28 (.34) & 0.83 (.33) & 0.94$^{\circ}$ (.33) \\
\bottomrule
\end{tabular}
\caption{Length inflation at each method's most relaxed tested setting, for all twelve pairs (first block: the primary pairs of Table~\ref{tab:length}): completion length relative to lossless ($\lambda$; $>1$ is inflation), with the accuracy at that setting in parentheses (MT-Bench: judge score out of 10). $^{\circ}$: the 95\% interval for $\lambda$ includes $1$ (primary pairs: Table~\ref{tab:cells}; other pairs: released tables).}
\label{tab:length-all}
\end{table*}

\begin{table*}[t]
\centering\scriptsize
\renewcommand{\arraystretch}{0.9}
\setlength{\tabcolsep}{3pt}
\begin{tabular}{@{}l ccccc ccccc ccccc@{}}
\toprule
Dataset & ment. & cactus & c-opt & r-fuzzy & c-tok & ment. & cactus & c-opt & r-fuzzy & c-tok & ment. & cactus & c-opt & r-fuzzy & c-tok \\
\midrule
 & \multicolumn{5}{c}{GPT-OSS-20B + EAGLE-3 (nebius)} & \multicolumn{5}{c}{Qwen3-8B + EAGLE-3 (RH)} & \multicolumn{5}{c}{GPT-OSS-20B + EAGLE-3 (RH)} \\
\cmidrule(lr){2-6}\cmidrule(lr){7-11}\cmidrule(lr){12-16}
GSM8K & \underline{0.83} & 0.99 & 1.07 & 1.14 & \underline{0.83} & \underline{0.93} & \underline{0.90} & 1.05 & \underline{0.76} & \underline{0.95} & 0.98 & \textbf{1.38} & 1.08 & \textbf{1.59} & \underline{0.83} \\
AIME24 & 1.18 & 1.07 & \textbf{1.70} & \textbf{1.37} & 1.22 & 0.94 & \underline{0.55} & 1.06 & 1.16 & 0.93 & 1.23 & \underline{0.46} & 1.18 & \textbf{1.67} & \textbf{1.33} \\
HumanEval & 0.97 & \textbf{1.12} & \textbf{1.29} & \textbf{1.16} & 0.93 & 0.99 & 0.96 & \textbf{1.36} & \textbf{1.33} & 1.04 & \textbf{1.30} & \textbf{1.80} & \textbf{1.43} & \textbf{2.70} & 1.02 \\
LiveCodeBench & \underline{0.89} & 0.94 & \textbf{1.16} & 1.11 & 0.98 & \underline{0.93} & \underline{0.72} & 1.03 & \underline{0.87} & \underline{0.94} & \textbf{1.34} & 1.08 & \textbf{1.38} & \textbf{1.28} & 1.06 \\
MT-Bench & \underline{0.75} & \underline{0.62} & \underline{0.76} & \underline{0.76} & \underline{0.88} & \underline{0.89} & \underline{0.64} & 0.94 & \underline{0.85} & 0.96 & 0.95 & \underline{0.74} & 0.99 & 0.93 & 1.03 \\
LongBench-v2 & 1.11 & 0.87 & \textbf{1.31} & 1.10 & 0.97 & 0.98 & \underline{0.80} & \underline{0.83} & 0.96 & 0.96 & 1.03 & \underline{0.58} & 0.92 & 0.97 & 1.02 \\
\addlinespace[1.5pt]
 & \multicolumn{5}{c}{Qwen3-8B + P-EAGLE} & \multicolumn{5}{c}{Qwen3-8B + DSpark} & \multicolumn{5}{c}{Qwen3-8B + Qwen3-0.6B} \\
\cmidrule(lr){2-6}\cmidrule(lr){7-11}\cmidrule(lr){12-16}
GSM8K & \underline{0.95} & \underline{0.94} & \underline{0.96} & \textbf{1.08} & 0.98 & \underline{0.80} & \underline{0.78} & \underline{0.83} & \underline{0.86} & \underline{0.82} & \underline{0.80} & \underline{0.75} & \underline{0.87} & \underline{0.72} & \underline{0.90} \\
AIME24 & 0.95 & 1.04 & 0.98 & \textbf{1.30} & 0.91 & 0.90 & 0.84 & 0.99 & 0.96 & 0.93 & \underline{0.78} & \underline{0.69} & \underline{0.80} & \underline{0.77} & \underline{0.88} \\
HumanEval & 0.97 & 0.98 & \textbf{1.08} & \textbf{1.42} & 0.97 & \underline{0.93} & \underline{0.93} & \textbf{1.09} & 1.01 & 0.95 & \underline{0.79} & \underline{0.76} & \underline{0.86} & \underline{0.71} & \underline{0.90} \\
LiveCodeBench & \underline{0.92} & \underline{0.94} & \underline{0.95} & \textbf{1.10} & 0.98 & \underline{0.86} & \underline{0.88} & \underline{0.89} & \underline{0.93} & \underline{0.87} & \underline{0.73} & \underline{0.69} & \underline{0.83} & \underline{0.68} & \underline{0.87} \\
MT-Bench & \underline{0.94} & \underline{0.84} & \underline{0.89} & 1.01 & 0.97 & \underline{0.79} & \underline{0.79} & \underline{0.77} & \underline{0.82} & \underline{0.84} & \underline{0.73} & \underline{0.60} & \underline{0.70} & \underline{0.58} & \underline{0.88} \\
LongBench-v2 & 0.95 & \underline{0.82} & 0.95 & 0.98 & 0.95 & \underline{0.82} & \underline{0.74} & \underline{0.67} & \underline{0.83} & \underline{0.81} & \underline{0.65} & \underline{0.54} & \underline{0.65} & \underline{0.56} & \underline{0.77} \\
\addlinespace[1.5pt]
 & \multicolumn{5}{c}{Qwen3-8B + Qwen3-1.7B} & \multicolumn{5}{c}{Llama-3.1-8B-I. + EAGLE-3} & \multicolumn{5}{c}{Llama-3.1-8B-I. + EAGLE-1} \\
\cmidrule(lr){2-6}\cmidrule(lr){7-11}\cmidrule(lr){12-16}
GSM8K & \underline{0.76} & \underline{0.76} & \underline{0.76} & \underline{0.75} & \underline{0.83} & 1.25 & \textbf{2.67} & 0.82 & 1.32 & \underline{0.63} & \textbf{1.39} & \textbf{3.08} & \textbf{1.35} & \textbf{3.18} & \underline{0.73} \\
AIME24 & \underline{0.66} & \underline{0.67} & \underline{0.74} & \underline{0.61} & \underline{0.77} & \textbf{1.70} & 1.08 & \underline{0.21} & 1.08 & \underline{0.30} & \textbf{1.45} & \textbf{1.64} & 0.80 & 0.76 & \underline{0.41} \\
HumanEval & \underline{0.75} & \underline{0.73} & \underline{0.74} & \underline{0.75} & \underline{0.82} & \textbf{1.37} & \textbf{8.47} & \textbf{1.34} & \textbf{5.16} & 0.89 & 1.00 & \textbf{4.98} & \textbf{3.33} & \textbf{1.59} & 0.94 \\
LiveCodeBench & \underline{0.70} & \underline{0.69} & \underline{0.72} & \underline{0.69} & \underline{0.80} & 0.94 & \textbf{6.21} & 1.55 & \textbf{2.24} & \underline{0.65} & \textbf{2.59} & \textbf{1.80} & \textbf{1.68} & \textbf{4.81} & \underline{0.69} \\
MT-Bench & \underline{0.67} & \underline{0.67} & \underline{0.67} & \underline{0.66} & \underline{0.77} & \textbf{1.59} & \textbf{2.12} & \underline{0.61} & \textbf{6.01} & \underline{0.67} & \textbf{1.79} & \textbf{3.22} & 0.93 & \textbf{4.01} & \underline{0.78} \\
LongBench-v2 & \underline{0.62} & \underline{0.61} & \underline{0.54} & \underline{0.65} & \underline{0.70} & \textbf{2.19} & \textbf{3.22} & \textbf{1.53} & \textbf{2.18} & 1.04 & 1.37 & \textbf{4.26} & 1.35 & 1.25 & \underline{0.62} \\
\addlinespace[1.5pt]
 & \multicolumn{5}{c}{Llama-3.1-8B-I. + Llama-3.2-1B} & \multicolumn{5}{c}{R1-Distill-Llama-8B + EAGLE-3} & \multicolumn{5}{c}{R1-Distill-Llama-8B + Llama-3.2-1B} \\
\cmidrule(lr){2-6}\cmidrule(lr){7-11}\cmidrule(lr){12-16}
GSM8K & \underline{0.68} & 1.06 & \underline{0.68} & 0.86 & \underline{0.74} & \underline{0.87} & 1.00 & \underline{0.89} & 1.11 & \underline{0.90} & \underline{0.87} & \textbf{1.37} & \textbf{2.30} & \textbf{1.99} & \underline{0.89} \\
AIME24 & \underline{0.26} & \underline{0.40} & \underline{0.23} & \underline{0.39} & 0.80 & \underline{0.80} & \underline{0.75} & \underline{0.72} & \underline{0.79} & 0.94 & \underline{0.76} & 0.98 & 0.99 & \underline{0.69} & 1.00 \\
HumanEval & \underline{0.83} & \underline{0.87} & \underline{0.90} & 0.90 & \underline{0.84} & 1.01 & 0.95 & 1.01 & \textbf{1.22} & 0.97 & 0.93 & \underline{0.80} & 0.97 & \textbf{1.26} & \underline{0.81} \\
LiveCodeBench & \underline{0.74} & 1.01 & \underline{0.71} & 0.93 & \underline{0.69} & 0.96 & \underline{0.44} & \underline{0.72} & 1.00 & \underline{0.92} & \underline{0.84} & \underline{0.50} & \underline{0.78} & \underline{0.74} & \underline{0.78} \\
MT-Bench & 0.94 & \textbf{1.55} & \underline{0.81} & 1.05 & \underline{0.83} & \underline{0.85} & \underline{0.67} & 0.89 & 1.18 & \underline{0.82} & \underline{0.75} & 0.96 & 0.99 & \textbf{1.19} & \underline{0.77} \\
LongBench-v2 & 0.77 & \textbf{1.58} & \underline{0.65} & 1.26 & \underline{0.56} & 1.01 & 0.99 & \underline{0.83} & 0.99 & 1.01 & \underline{0.68} & 1.11 & \textbf{2.05} & \underline{0.73} & \underline{0.72} \\
\bottomrule
\end{tabular}
\caption{Verifier rounds relative to lossless ($R/R^{*}$) at each method's most relaxed tested setting, for all twelve pairs (first two: the primary pairs of Table~\ref{tab:rounds}; RH: RedHatAI; Llama-3.1-8B-I.: Llama-3.1-8B-Instruct; ment., c-: mentored, casc-). Bold: the 95\% interval for $R/R^{*}$ lies entirely above $1$ (more rounds than lossless); underline: entirely below $1$; intervals are in the released tables.}
\label{tab:rounds-all}
\end{table*}

\begin{table*}[t]
\centering\scriptsize
\setlength{\tabcolsep}{2pt}
\begin{tabular}{@{}l rrr rrr rrr rrr rrr rrr@{}}
\toprule
& \multicolumn{3}{c}{GSM8K} & \multicolumn{3}{c}{AIME24} & \multicolumn{3}{c}{HumanEval} & \multicolumn{3}{c}{LiveCodeBench} & \multicolumn{3}{c}{MT-Bench} & \multicolumn{3}{c}{LongBench-v2} \\
\cmidrule(lr){2-4}\cmidrule(lr){5-7}\cmidrule(lr){8-10}\cmidrule(lr){11-13}\cmidrule(lr){14-16}\cmidrule(lr){17-19}
Target, drafter & $\laccept^{*}$ & $\Lbar^{*}$ & acc. & $\laccept^{*}$ & $\Lbar^{*}$ & acc. & $\laccept^{*}$ & $\Lbar^{*}$ & acc. & $\laccept^{*}$ & $\Lbar^{*}$ & acc. & $\laccept^{*}$ & $\Lbar^{*}$ & score & $\laccept^{*}$ & $\Lbar^{*}$ & acc. \\
\midrule
\multicolumn{19}{@{}l}{GPT-OSS-20B} \\
\quad EAGLE-3 (RedHatAI) & 2.13 & 353 & .97 & 1.34 & 9\,650 & .77 & 2.04 & 911 & .98 & 1.57 & 3\,445 & .83 & 1.82 & 1\,157 & 7.31 & 0.05 & 1\,403 & .54 \\
\multicolumn{19}{@{}l}{Qwen3-8B} \\
\quad P-EAGLE & 2.16 & 1\,303 & .81 & 2.26 & 18\,758 & .73 & 2.05 & 3\,863 & .85 & 1.84 & 8\,072 & .73 & 1.57 & 2\,112 & 6.96 & 1.38 & 3\,023 & .55 \\
\quad DSpark & 2.77 & 1\,335 & .75 & 2.51 & 19\,550 & .70 & 2.50 & 3\,756 & .84 & 2.31 & 8\,063 & .74 & 2.12 & 2\,133 & 6.74 & 1.88 & 2\,948 & .47 \\
\quad Qwen3-0.6B & 3.16 & 1\,310 & .77 & 3.58 & 17\,807 & .67 & 2.90 & 3\,814 & .84 & 3.01 & 8\,076 & .72 & 2.38 & 1\,997 & 6.80 & 2.51 & 2\,989 & .46 \\
\quad Qwen3-1.7B & 3.82 & 1\,316 & .79 & 4.15 & 19\,469 & .73 & 3.71 & 3\,876 & .84 & 3.80 & 8\,098 & .69 & 3.04 & 2\,098 & 6.80 & 3.07 & 2\,896 & .45 \\
\multicolumn{19}{@{}l}{Llama-3.1-8B-Instruct} \\
\quad EAGLE-3 & 2.34 & 290 & .81 & 0.77 & 3\,661 & .03 & 2.70 & 547 & .56 & 2.05 & 770 & .14 & 2.08 & 472 & 5.85 & 0.06 & 525 & .37 \\
\quad EAGLE-1 & 1.57 & 279 & .79 & 0.77 & 4\,842 & .03 & 1.76 & 529 & .58 & 1.42 & 830 & .19 & 1.32 & 468 & 5.95 & 0.92 & 705 & .35 \\
\quad Llama-3.2-1B & 4.04 & 269 & .78 & 3.56 & 5\,380 & .00 & 3.97 & 544 & .55 & 3.48 & 764 & .16 & 3.26 & 430 & 5.53 & 3.11 & 665 & .37 \\
\multicolumn{19}{@{}l}{R1-Distill-Llama-8B} \\
\quad EAGLE-3 & 3.07 & 623 & .77 & 0.60 & 12\,527 & .27 & 1.89 & 3\,739 & .86 & 0.77 & 8\,882 & .58 & 2.13 & 1\,384 & 5.40 & 0.03 & 1\,911 & .29 \\
\quad Llama-3.2-1B & 2.71 & 440 & .71 & 1.83 & 11\,025 & .43 & 1.69 & 3\,857 & .89 & 1.56 & 8\,742 & .56 & 1.64 & 1\,373 & 5.30 & 1.62 & 1\,802 & .34 \\
\bottomrule
\end{tabular}
\caption{Lossless references of the ten pairs beyond the primary ones (primary pairs: Table~\ref{tab:strict}; same cases and budgets), seed 0: mean accepted length $\laccept^{*}$, mean completion tokens $\Lbar^{*}$, and accuracy (MT-Bench: judge score out of 10).}
\label{tab:strict-all}
\end{table*}

\begin{table*}[t]
\centering\scriptsize
\setlength{\tabcolsep}{4pt}
\begin{tabular}{@{}l r rrrr rrrr@{}}
\toprule
& & \multicolumn{4}{c}{gentlest tested setting} & \multicolumn{4}{c}{most relaxed tested setting} \\
\cmidrule(lr){3-6}\cmidrule(lr){7-10}
Pair & same & $\lambda$ & $R/R^{*}$ & $R/R^{*}>1$ & acc.\ loss & $\lambda$ & $R/R^{*}$ & $R/R^{*}>1$ & acc.\ loss \\
\midrule
GPT-OSS-20B + EAGLE-3 (nebius) & 0 & 1.07 & 0.98 & 9 & 14 & 1.57 & 1.03 & 15 & 22 \\
GPT-OSS-20B + EAGLE-3 (RedHatAI) & 14 & 1.23 & 1.09 & 22 & 15 & 1.35 & 1.07 & 20 & 17 \\
Qwen3-8B + EAGLE-3 (RedHatAI) & 0 & 1.04 & 0.99 & 12 & 13 & 1.08 & 0.94 & 7 & 18 \\
Qwen3-8B + DSpark & 8 & 1.01 & 0.89 & 1 & 12 & 1.06 & 0.85 & 2 & 18 \\
Llama-3.1-8B-Instruct + EAGLE-3 & 7 & 1.84 & 1.57 & 24 & 18 & 1.69 & 1.35 & 21 & 18 \\
Llama-3.1-8B-Instruct + EAGLE-1 & 12 & 1.23 & 1.13 & 18 & 18 & 2.42 & 1.42 & 21 & 19 \\
R1-Distill-Llama-8B + EAGLE-3 & 14 & 1.07 & 0.95 & 11 & 15 & 1.11 & 0.93 & 8 & 16 \\
\midrule
All (210 cells) & 55 & 1.07 & 0.99 & 97 & 105 & 1.22 & 0.97 & 94 & 128 \\
\bottomrule
\end{tabular}
\caption{Gentlest against most relaxed tested setting on the seven pairs run with the matched protocol (30 cells each; the settings with the lowest and the highest measured $\laccept$): median $\lambda$ and median $R/R^{*}$ over the pair's cells, cells that need more rounds than lossless ($R/R^{*}>1$), and graded cells two or more points below lossless (acc.\ loss, of 25; All: of 175). same: cells whose two settings are the same run.}
\label{tab:gentle}
\end{table*}

\section{SPEED-Bench}
\label{app:speedbench}

SPEED-Bench \citep{abramovich2026speedbench} has 880 prompts in eleven categories, built to measure speculative decoding across request types, and no accuracy grading; it is the benchmark closest to a serving workload, and \citet{xia2026practical} do not use it. We run the five rules at their loosest settings and lossless verification on both primary pairs (seed 0, budget 8\,192 tokens) on the 672 prompts not drawn from the gated HLE set: eight categories are complete (80 prompts each), and Humanities, Math, and STEM keep 8, 18, and 6 prompts. Lossless completions are short here, and the acceptance gain of every rule but one exceeds its inflation, so rounds fall (Table~\ref{tab:speed}); the exception is \texttt{spec\_casc\_opt} on Qwen3-8B, which inflates by $1.40\times$ against a gain of $1.32\times$ and needs $1.02\times$ the rounds. Equation~\eqref{eq:netwin} predicts each of the ten round ratios within $7\%$ and its direction in all ten. Per category (released tables), the round ratio's interval lies below $1$ in 31 of the 55 (rule, category) cells on GPT-OSS-20B and in 24 on Qwen3-8B, and above $1$ in 3 and 1, all four with $\lambda \ge 1.71$. On GPT-OSS-20B the rounds verdict and the time verdict agree in every cell; on Qwen3-8B the time ratios carry the third server's node effect (Appendix~\ref{app:seeds}) and are reported over same-node pairs where possible. Quality is not graded on this benchmark, so the rules that save the most rounds here, \texttt{cactus} above all, are the ones that lose the most accuracy and judged quality in Section~\ref{sec:accuracy}.

\begin{table}[t]
\centering\scriptsize
\setlength{\tabcolsep}{2.5pt}
\begin{tabular}{@{}l rrrr rrrr@{}}
\toprule
& \multicolumn{4}{c}{GPT-OSS-20B} & \multicolumn{4}{c}{Qwen3-8B} \\
\cmidrule(lr){2-5}\cmidrule(lr){6-9}
Rule & $\lambda$ & gain & $R/R^{*}$ & $T$ & $\lambda$ & gain & $R/R^{*}$ & $T$ \\
\midrule
\texttt{mentored\_dec} & \textbf{1.15} & 1.44 & \underline{0.76} & 0.75 & \textbf{1.07} & 1.14 & \underline{0.93} & 0.91$^{\dagger}$ \\
\texttt{cactus} & \textbf{1.26} & 1.89 & \underline{0.63} & 0.62 & \textbf{1.35} & 1.91 & \underline{0.67} & 0.68$^{\ddagger}$ \\
\texttt{spec\_casc\_opt} & \textbf{1.23} & 1.41 & \underline{0.82} & 0.82 & \textbf{1.40} & 1.32 & 1.02 & 1.06$^{\dagger}$ \\
\texttt{r\_fuzzy} & \textbf{1.20} & 1.52 & \underline{0.77} & 0.75 & \textbf{1.20} & 1.28 & \underline{0.95} & 0.96 \\
\texttt{spec\_casc\_tok} & \textbf{1.09} & 1.13 & \underline{0.95} & 0.92 & 1.01 & 1.05 & \underline{0.95} & 0.95$^{\dagger}$ \\
\bottomrule
\end{tabular}
\caption{SPEED-Bench over all 672 prompts, loosest settings, seed 0 (GPT-OSS-20B on the second server; Qwen3-8B's first 40 prompts per arm on the second, the rest on the third). Lossless: 1\,266 and 2\,332 tokens on average, 1\% and 5\% cut off. $\lambda$, $R/R^{*}$, $T$: ratios of completion length, verifier rounds, and per-request time to lossless; gain: $(\laccept+1)/(\laccept^{*}+1)$. Bold (underline): 95\% paired interval above (below) $1$. $^{\dagger}$: over the prompt pairs that share a node; $^{\ddagger}$: fewer than ten do, all-pair ratio.}
\label{tab:speed}
\end{table}

\section{Forcing the Reasoning Trace to End}
\label{app:buys}

\paragraph{Guard search.} A coding agent edited one function that returns, for each verification round, the draft positions to verify losslessly; at a marked position \texttt{spec\_casc\_tok} reduces to its $\alpha \to -\infty$ limit, which is lossless verification. After each edit the agent replayed eight AIME24 cases on GPT-OSS-20B ($\alpha = 0.3$, one fresh server per case) and kept the edit if mean completion length fell without losing more than one correct answer. Over eleven iterations, the best guard (lossless verification for eight positions after clusters of hesitation and discourse markers, and for whole rounds after 14\,000 tokens without a final channel) cut the eight cases from 13\,364 to 10\,534 tokens at 6 of 8 correct; on all 30 cases it moved the mean from 9\,270 to 10\,363 tokens and accuracy from 22 to 23. On the eight cases, two runs that reached the 32\,768-token budget without closing their reasoning trace held $61\%$ of all output tokens. A guard judges one draft token at a time and cannot see whether a run will ever close its trace, so Section~\ref{sec:buys} acts on the run instead.

\paragraph{Forcing.} For Qwen3-8B, a hook in vLLM's sampler writes \texttt{</think>} at the first draft position of a round once the request has generated the threshold number of tokens without closing its trace; the hook does not depend on the acceptance rule, so one hook serves all five rules. For GPT-OSS-20B, a patch to each rule injects the tokens that open the final channel. The threshold is the budget minus the 95th percentile of the answer length after the closing token in the pair's lossless runs (characters converted at 3.25 per token); it was chosen after a first pass at $0.9$ of the budget left too little room for the answer, so it was not fixed in advance. The forced runs are the relaxed runs that reached the budget or passed the threshold without closing their trace; the hook is inactive below the threshold, so every other run is unchanged. The baselines of \texttt{spec\_casc\_tok} and \texttt{mentored\_dec} were rerun in one session; the other three rules use the runs of the sweep and its pilot (Section~\ref{sec:protocol}). A run cut off before closing its reasoning trace is scored wrong. Qwen3-8B reruns reproduce earlier runs (127 of 127 LongBench-v2 runs of \texttt{spec\_casc\_tok} below the threshold keep their exact token count); GPT-OSS-20B reruns on a server reused across requests do not (59 of 146), so its rescued and lost counts include trajectory noise, and the fresh-server checks below are its clean measurement.

\paragraph{Forcing results.} GPT-OSS-20B, at its recommended sampler, gains much less than Qwen3-8B (33 rescued and 15 lost of 177 runs). The gentle rules recover most: on Qwen3-8B, $66\%$ and $68\%$ of forced \texttt{mentored\_dec} and \texttt{spec\_casc\_tok} runs end correct, against $28\%$ to $38\%$ for the three aggressive rules (Tables~\ref{tab:force} and~\ref{tab:force-all}). Forcing does not always end the run: 515 forced runs still reach the budget ($23\%$ on GSM8K, $50\%$ to $68\%$ on the other datasets), as the model resumes reasoning or rewrites its solution after the forced token. Two paired checks on fresh servers, where reruns are reproducible byte for byte, show that the intervention is surgical: on GPT-OSS-20B AIME24, at thresholds of 28\,000 and 30\,000 tokens, it leaves 27 and 23 of 30 outputs unchanged, removes every run that reached the budget (3 and 7), and keeps accuracy at 22 of 30 for \texttt{spec\_casc\_tok} ($\alpha = 0.3$) while raising it from 16 to 19 for \texttt{mentored\_dec}, with no correct answer lost.

\paragraph{Answer timing.} For each output we find the first occurrence of the reference answer as a standalone number (thousands separators allowed) and convert its character offset to tokens in proportion to the output's token count. Cases whose answer appears in the prompt are excluded (12 of 150 on GSM8K, none on AIME24). A pair is a case and a rule for which both arms answer correctly and neither is cut off; \texttt{spec\_casc\_tok} is left out because its completions do not grow. Intervals are 95\% bootstrap intervals over cases (5\,000 resamples; a resampled case keeps all its rules). The first occurrence is a proxy: an intermediate result can equal the answer. In one Qwen3-8B case under \texttt{spec\_casc\_opt}, both arms write the answer near token 300; lossless verification answers within 1\,292 tokens, while the relaxed run writes the answer 23 more times and is cut off at 2\,048 tokens with no answer.

\begin{table*}[t]
\centering\scriptsize
\setlength{\tabcolsep}{4.5pt}
\begin{tabular}{@{}l rr@{\,$\to$\,}lrrr rr@{\,$\to$\,}lrrr@{}}
\toprule
 & \multicolumn{6}{c}{Qwen3-8B} & \multicolumn{6}{c}{GPT-OSS-20B} \\
\cmidrule(lr){2-7}\cmidrule(lr){8-13}
Dataset, rule & forced & \multicolumn{2}{c}{correct} & $+$ & $-$ & cut off & forced & \multicolumn{2}{c}{correct} & $+$ & $-$ & cut off \\
\midrule
\multicolumn{13}{@{}l}{GSM8K} \\
\quad \texttt{mentored\_dec} & 44 & 6 & 38 & 32 & 0 & 4 & 6 & 2 & 2 & 0 & 0 & 3 \\
\quad \texttt{cactus} & 56 & 15 & 45 & 30 & 0 & 11 & 9 & 1 & 2 & 1 & 0 & 6 \\
\quad \texttt{spec\_casc\_opt} & 102 & 18 & 60 & 44 & 2 & 27 & 7 & 1 & 2 & 1 & 0 & 4 \\
\quad \texttt{r\_fuzzy} & 62 & 8 & 47 & 39 & 0 & 13 & 8 & 0 & 3 & 3 & 0 & 3 \\
\quad \texttt{spec\_casc\_tok} & 43 & 9 & 35 & 26 & 0 & 6 & 2 & 0 & 1 & 1 & 0 & 1 \\
\multicolumn{13}{@{}l}{AIME24} \\
\quad \texttt{mentored\_dec} & 6 & 0 & 1 & 1 & 0 & 3 & 6 & 0 & 0 & 0 & 0 & 3 \\
\quad \texttt{cactus} & 17 & 1 & 1 & 0 & 0 & 14 & 2 & 0 & 0 & 0 & 0 & 0 \\
\quad \texttt{spec\_casc\_opt} & 19 & 0 & 1 & 1 & 0 & 17 & 13 & 0 & 4 & 4 & 0 & 9 \\
\quad \texttt{r\_fuzzy} & 13 & 1 & 2 & 1 & 0 & 9 & 9 & 1 & 1 & 1 & 1 & 5 \\
\quad \texttt{spec\_casc\_tok} & 5 & 0 & 1 & 1 & 0 & 2 & 6 & 2 & 2 & 2 & 2 & 3 \\
\multicolumn{13}{@{}l}{HumanEval} \\
\quad \texttt{mentored\_dec} & 22 & 2 & 12 & 10 & 0 & 5 & 1 & 0 & 0 & 0 & 0 & 0 \\
\quad \texttt{cactus} & 32 & 1 & 10 & 9 & 0 & 16 & 3 & 0 & 0 & 0 & 0 & 2 \\
\quad \texttt{spec\_casc\_opt} & 67 & 7 & 18 & 11 & 0 & 39 & 4 & 1 & 0 & 0 & 1 & 0 \\
\quad \texttt{r\_fuzzy} & 65 & 4 & 3 & 0 & 1 & 38 & 1 & 1 & 0 & 0 & 1 & 0 \\
\quad \texttt{spec\_casc\_tok} & 19 & 2 & 9 & 7 & 0 & 8 & \multicolumn{6}{c}{no forced runs} \\
\multicolumn{13}{@{}l}{LiveCodeBench} \\
\quad \texttt{mentored\_dec} & 33 & 2 & 25 & 23 & 0 & 10 & 10 & 1 & 3 & 3 & 1 & 2 \\
\quad \texttt{cactus} & 51 & 3 & 18 & 16 & 1 & 33 & 1 & 0 & 0 & 0 & 0 & 0 \\
\quad \texttt{spec\_casc\_opt} & 58 & 3 & 14 & 12 & 1 & 30 & 21 & 1 & 6 & 5 & 0 & 10 \\
\quad \texttt{r\_fuzzy} & 55 & 0 & 0 & 0 & 0 & 42 & 20 & 0 & 4 & 4 & 0 & 10 \\
\quad \texttt{spec\_casc\_tok} & 29 & 3 & 24 & 21 & 0 & 5 & 10 & 5 & 3 & 1 & 3 & 1 \\
\multicolumn{13}{@{}l}{LongBench-v2} \\
\quad \texttt{mentored\_dec} & 10 & 2 & 0 & 0 & 2 & 2 & 4 & 3 & 1 & 0 & 2 & 1 \\
\quad \texttt{cactus} & 89 & 8 & 19 & 11 & 0 & 75 & 3 & 0 & 0 & 0 & 0 & 0 \\
\quad \texttt{spec\_casc\_opt} & 31 & 3 & 5 & 2 & 0 & 24 & 22 & 5 & 4 & 3 & 4 & 7 \\
\quad \texttt{r\_fuzzy} & 19 & 3 & 7 & 4 & 0 & 6 & 8 & 0 & 3 & 3 & 0 & 3 \\
\quad \texttt{spec\_casc\_tok} & 8 & 1 & 2 & 1 & 0 & 3 & 1 & 0 & 1 & 1 & 0 & 0 \\
\midrule
Total & 955 & 102 & 397 & 302 & 7 & 442 & 177 & 24 & 42 & 33 & 15 & 73 \\
\bottomrule
\end{tabular}
\caption{Table~\ref{tab:force} by dataset and rule, each rule at its loosest tested setting. Columns as in Table~\ref{tab:force}, plus \emph{cut off}: forced runs that still end at the budget. Thresholds in tokens (Qwen3-8B / GPT-OSS-20B): GSM8K 1\,684 / 2\,002; AIME24 31\,260 / 31\,605; HumanEval 8\,150 / 8\,497; LiveCodeBench 11\,004 / 9\,294; LongBench-v2 6\,894 / 8\,133.}
\label{tab:force-all}
\end{table*}

\begin{table*}[t]
\centering\scriptsize
\setlength{\tabcolsep}{5pt}
\begin{tabular}{@{}ll r rr rr r rr@{}}
\toprule
 & & & \multicolumn{2}{c}{First answer (token)} & \multicolumn{2}{c}{Length (tokens)} & Extra tokens & \multicolumn{2}{c}{Cut-off runs with the answer} \\
\cmidrule(lr){4-5}\cmidrule(lr){6-7}\cmidrule(lr){9-10}
Target & Dataset & Pairs & lossless & relaxed & lossless & relaxed & after answer (\%) & relaxed & lossless \\
\midrule
GPT-OSS-20B & GSM8K & 486 & 102 & 126 & 183 & 264 & 50 [37, 59] & 15/26 & 0/2 \\
Qwen3-8B & GSM8K & 304 & 325 & 350 & 877 & 989 & 82 [65, 96] & 147/201 & 28/33 \\
GPT-OSS-20B & AIME24 & 36 & 1\,721 & 3\,865 & 3\,822 & 7\,020 & 41 [28, 54] & 8/33 & 1/2 \\
Qwen3-8B & AIME24 & 33 & 4\,221 & 6\,031 & 11\,160 & 14\,230 & $-16$ [$-177$, 49] & 9/53 & 1/5 \\
\bottomrule
\end{tabular}
\caption{Where the extra tokens sit (Section~\ref{sec:buys}): the four rules other than \texttt{spec\_casc\_tok}, each at its loosest setting. Pairs: (case, rule) pairs in which both arms answer correctly and neither is cut off; first answer and length: medians over pairs; extra tokens after answer: share of the summed extra length that falls after the point where each run first writes the answer, with a 95\% bootstrap interval over cases (negative: the part after the answer shrinks while the part before it grows); cut-off runs with the answer: runs that end at the budget without a correct answer, and how many of them contain it.}
\label{tab:timing}
\end{table*}

\end{document}